# DERM-3R: A Resource-Efficient Multimodal Agents Framework for Dermatologic Diagnosis and Treatment in Real-World Clinical Settings

Jirui Dai[4,一], Zhendong Wang[2,*], Chongjing Wang[9,*], Yusha Bai[5], Ziwen Chen[6], Jihao Gu[7], Zhou Zhang[8], Changyong Luo[3,†], Haibing Lan[2,†], Zhi Liu[1,二]

## Abstract

Dermatology diseases impose a substantial and growing global health burden, affecting billions of individuals worldwide and significantly impairing people's life. Although modern dermatologic therapies enable rapid control of acute manifestations, their long-term effectiveness is limited by single-target treatment paradigms, recurrent disease courses, and insufficient consideration of systemic comorbidities. Traditional Chinese medicine (TCM) offers a complementary, holistic approach to dermatologic care through syndrome differentiation and individualized treatment strategies. However, its clinical practice is challenged by non-standardized knowledge system, multimodal records with missing essential items, and difficulties in scaling expert-driven diagnostic and therapeutic reasoning. To handle these challenges, we propose DERM-3R, a resource-efficient multimodal agents framework designed to model dermatologic diagnosis and treatment knowledge and skills in TCM under limited data and computational resources. By analyzing real-world clinical settings, we reformulate dermatologic decision-making through task decomposition and redefinition, structuring the process into three essential issues: fine-grained lesion recognition, multi-view lesion representation with specialist-level pathogenesis modeling, and holistic clinical reasoning for syndrome differentiation and treatment planning. DERM-3R consists of three collaborative agents—DERM-Rec, DERM-Rep, and DERM-Reason—each addressing a specific component of the clinical reasoning process. Built upon a lightweight multimodal large language model and fine-tuned by partial-parameter finetuning on 103 real-world TCM psoriasis cases, DERM-3R demonstrates strong performance across multiple dermatologic reasoning tasks. Comprehensive evaluations using traditional automatic metrics, LLM-as-a-Judge assessments, and human doctors show that, despite extremely limited training data and parameter updates, DERM-3R achieves performance comparable to even surpassing that of hundred-billion-parameter general-purpose multimodal models, including GPT-5.1 and Gemini-3-Flash. Our results indicate that structured, domain-aware multi-agent modeling provides an effective and alternative way to brute-force scaling for complex clinical tasks, offering a practical and scalable paradigm for multimodal AI applications in dermatology and integrative medicine.

## Introduction

The skin, as the largest organ covering the human body, plays essential roles in protection, thermoregulation, sensory perception, and immune defense. Pathological alterations of the skin impose a substantial burden on public health. As of 2021, the global incidence of skin diseases reached approximately 4.69 billion cases, ranking seventh among all diseases in terms of years

一 The first three authors are the co-first authors.

二 Corresponding authors: Zhi Liu (zhiliu@njucm.edu.cn), Changyong Luo (Bdf01344@bucm.edu.cn), Haibing Lan (13621362615@126.com)

lived with disability (YLDs), indicating that nearly six out of every ten individuals worldwide are affected by skin diseases[1]. At the 78th World Health Assembly concluded on May 27, 2025, Member States of the World Health Organization (WHO) formally adopted a resolution recognizing skin diseases as a global public health priority[2]. The burden of skin diseases is primarily reflected in their profound impact on appearance and social functioning, frequently leading to anxiety, depression, and other psychological disorders; persistent pruritus or pain associated with many dermatologic conditions significantly reduces quality of life; moreover, certain skin diseases represent cutaneous manifestations of systemic disorders or are accompanied by comorbidities such as joint damage and cardiovascular disease.

Modern dermatologic practice is largely guided by clinical guidelines and relies on therapeutic approaches including corticosteroids, immunosuppressants, biologics, and small-molecule targeted agents. While these treatments offer advantages in standardized diagnosis and rapid control of acute manifestations, their limitations have become increasingly evident[3]. Current therapies often focus on single targets, where treating one condition may induce new complications. Long-term use of immunosuppressants or biologics is frequently associated with diminished efficacy and limited prevention of disease recurrence[4,5]. Furthermore, insufficient attention is paid to comorbidities and the overall systemic condition of patients[6]. In response to these challenges, growing attention has been directed toward the role of holistic microenvironmental imbalance in dermatologic pathology, alongside increasing interest in natural plant- and animal-derived medicines, with promising therapeutic outcomes reported[7,8].

As a representative medical system utilizing natural and processed botanical and zoological medicines, traditional Chinese medicine (TCM) has been applied for over a millennium and is supported by a systematic theoretical framework and extensive clinical practice. In TCM, skin diseases are regarded as external manifestations of internal imbalances in Yin–Yang and dysfunctions of the Zang–Fu organs. Diagnosis is based on comprehensive syndrome differentiation, integrating lesion characteristics with systemic symptoms, followed by individualized treatment strategies under the principle of “one patient, one prescription.” This approach offers distinctive advantages and substantial potential. **First**, multi-component herbal formulas exert multi-target effects across multiple biological pathways, regulating immune, neural, and endocrine systems in addition to addressing local lesions, thereby restoring systemic homeostasis, accelerating recovery, and reducing recurrence rates[9-11]. **Second**, in dermatologic practice, TCM not only alleviates hallmark symptoms such as pruritus but also improves sleep quality, emotional state, and overall well-being[12,13]. Despite these strengths, the methodology and practices of TCM dermatology faces intrinsic challenges. Its highly individualized nature results in heterogeneous and non-standardized data, posing significant difficulties for knowledge formalization. High-quality dermatologic case accumulation is challenging, with frequent incompleteness of records and substantial loss of lesion-centered image data critical for syndrome differentiation. Moreover, TCM-based dermatologic diagnosis requires long-term, iterative integration of theoretical knowledge and clinical experience, constituting a complex knowledge engineering problem with strongly coupled multi-task learning characteristics. The lack of standardization in TCM knowledge systems not only increases the learning burden for practitioners but also contributes to the prolonged and demanding training process required for cultivating high-level TCM clinicians.

In light of these challenges, large language models (LLMs), with their capacity to learn complex multi-task associations from large-scale data and to internalize structured reasoning patterns, offer a novel technical paradigm for addressing issues such as implicit experiential knowledge, fragmented representations, and personalized decision-making in TCM[14]. However, current researches on LLMs have excessively emphasized large-scale parameterization and trillion-token datasets to demonstrate general-purpose capabilities. **This paradigm results in high energy consumption, prohibitive deployment costs, and severe hallucination issues in specialized medical domains**, ultimately limiting their practical applicability in clinical medicine, including TCM. **Furthermore**, in TCM dermatology, visual information derived from skin lesions—such as color, morphology, texture, distribution, and qualitative characteristics—constitutes a core basis for syndrome differentiation and treatment decisions[15,16]. **In addition to visual cues**, seemingly unrelated systemic symptoms expressed in textual form also play a critical role in TCM diagnostic reasoning. Consequently, effective modeling of TCM dermatologic diagnosis requires joint reasoning over both visual and textual modalities[15]. However, due to incomplete case records, severe loss of image–text correspondence, and non-standardized terminology, both existing unimodal and multimodal large models are inherently limited under conventional research paradigms[17].

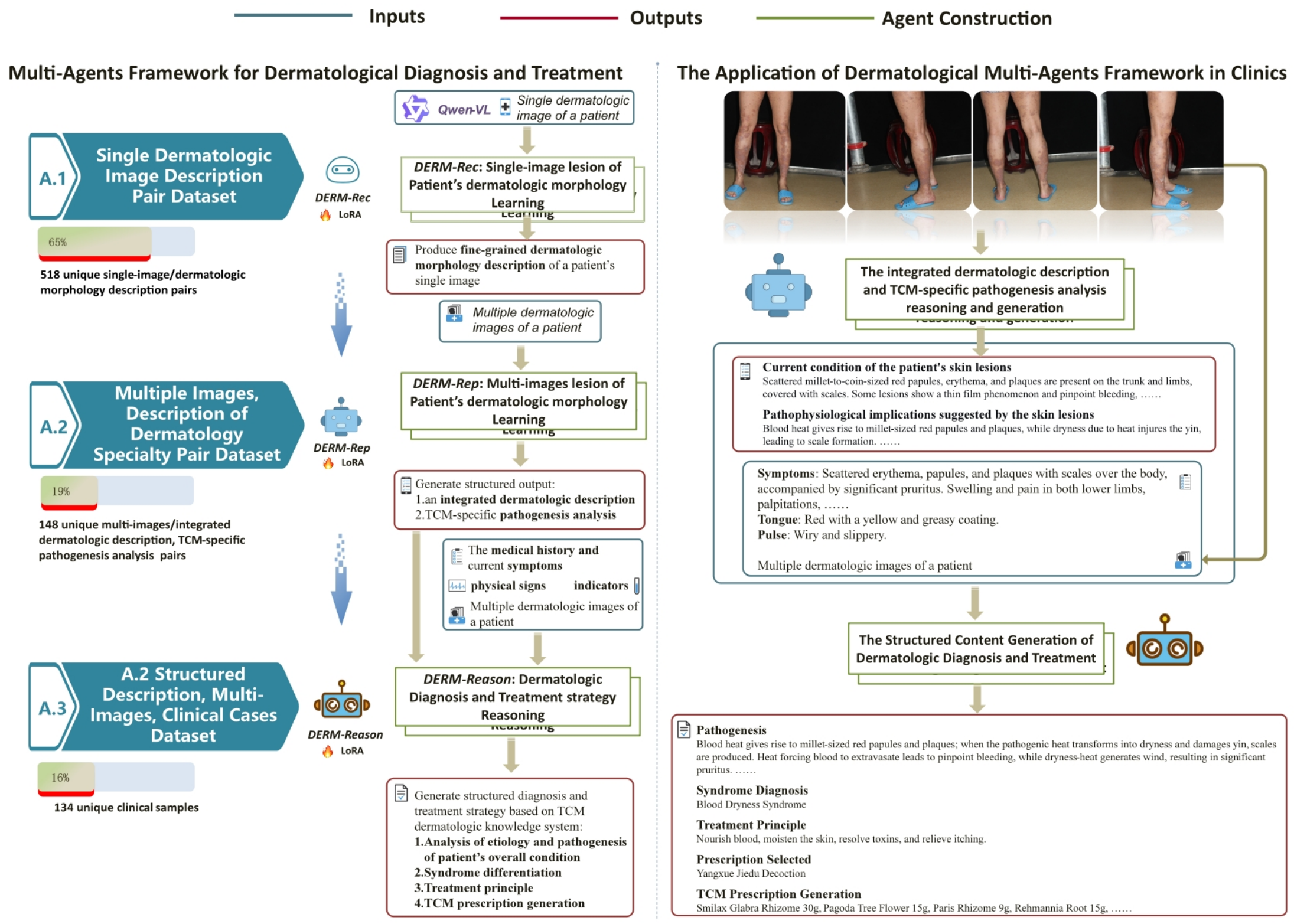

Fig. 1. The DERM-3R framework we proposed to reform the clinical diagnosis and treatment tasks of dermatologic patient through multi-agent collaboration. We analyze and reform it as three multi-modal agents to alleviate three key aspects of TCM dermatologic diagnosis and treatment process: fine-grained lesion perception and recognition, multi-view lesion aggregation with specialist pathogenesis modeling, and holistic syndrome-based diagnostic and therapeutic reasoning. As shown in the left part, The DERM-3R framework comprises three collaborative agents. The Dermatologic Recognition Agent (DERM-Rec) is responsible for recognizing lesions from single-image inputs, performing morphological analysis, and generating visual-semantic representations. The Dermatologic Representation Agent (DERM-Rep) aggregates multiple lesion images to develop a comprehensive understanding of the patient's overall dermatological condition, along with the associated TCM pathogenesis,

producing detailed specialist descriptions and pathogenesis models. Finally, the Multimodal Clinical Reasoning Agent (DERM-Reason) integrates the outputs from DERM-Rep with multi-view lesion images, the patient's medical history, and current symptoms, enabling it to conduct thorough reasoning. This agent produces a full analysis of pathogenesis, syndrome differentiation, treatment strategies, and prescription recommendations. In right part, we display the inference process when deploying DERM-3R in the real-world clinical scenario.

To address these clinical and methodological challenges, we analyzed real-world TCM dermatologic clinical practice and reformulated the problem through task decomposition and redefinition. Building upon our prior work with the Tianyi[18], we propose a multi-agent framework, **DERM-3R**, as illustrated in Fig. 1. It employs coordinated multi-agent modeling with lightweight multimodal large models to address three key aspects of TCM dermatologic diagnosis and treatment: fine-grained lesion perception and recognition, multi-view lesion aggregation with specialist pathogenesis modeling, and holistic syndrome-based diagnostic and therapeutic reasoning. Through this design, DERM-3R effectively addresses challenges including complex multi-task modeling, scarcity of complete clinical samples, and limited computational resources deployment.

DERM-3R consists of three collaboratively agents. The **Dermatologic Recognition Agent (DERM-Rec)** models single-image lesion perception, morphological recognition, and visual–semantic representation. The **Dermatologic Representation Agent (DERM-Rep)** aggregates multi-view lesion images to form an integrated understanding of the patient's overall dermatologic condition and its corresponding TCM pathogenesis, generating structured specialist descriptions and pathogenesis representations. The **Multimodal Clinical Reasoning Agent (DERM-Reason)** integrates outputs from DERM-Rep with multi-view lesion images, patient medical history, and current symptoms to perform comprehensive reasoning, generating overall pathogenesis analysis, syndrome differentiation, treatment principles, and prescription recommendations. All agents is built upon the Qwen2.5-VL-7B[19] multimodal large model, integrated with the Tianyi language model and fine-tuned using LoRA[20] on 103 real-world TCM dermatologic cases of psoriasis. Each case has multiple clinical visits. Results from both automated evaluations and expert assessments demonstrate that, under limited computational resources and extremely small sample scale, DERM-3R achieves performance comparable even surpassing that of hundred-billion-parameter general-purpose multimodal models, including GPT-5.1 and Gemini-3-Flash.

The main contributions of this study are fourfold:

(1) We introduce DERM-3R, a clinically grounded multi-agent collaborative framework for dermatologic diagnosis and treatment. **It reformulates complex dermatologic decision-making as a staged multimodal reasoning process and enables effective coordination among specialized agents under real-world clinical constraints**.

(2) **We propose a task decomposition and redefinition paradigm derived from dermatologic clinical practice**, systematically translating expert-driven diagnostic workflows into AI-compatible modeling objectives, thereby providing a generalizable methodology for structured AI modeling in dermatologic and related medical specialties.

(3) Through task-specific sample reconstruction and staged- and joint-fintuning, we demonstrate that **high-quality multimodal modeling and reasoning abilities can be achieved with extremely limited annotated data and lightweight model parameters, offering practical evidence that data- and compute-efficient AI systems can support complex clinical tasks**.

(4) We empirically show that **lightweight multimodal models trained within the DERM-3R framework can outperform hundred-billion-parameter general-purpose multimodal models on challenging clinical multi-task settings**, highlighting the effectiveness of structured, domain-aware modeling over brute-force scaling in medical AI applications.

## 2. Method

### 2.1 Task Decomposition and Redefinition

The core of TCM–based dermatologic treatment lies in physicians' systematic analysis of dermatologic manifestations, medical history, and accompanying symptoms, grounded in established TCM theories of dermatology and accumulated clinical experience. Through such process, clinicians form a holistic understanding of the patient's current disease state and subsequently determine appropriate diagnostic conclusions and treatment strategies. This clinical process is not a simple mapping between symptoms and therapies, but rather a comprehensive medical activity characterized by holistic thinking, syndrome differentiation, and dynamic judgment.

In routine clinical practice, **first**, physicians conduct detailed observations of dermatologic manifestations, accurately identifying lesion morphology, distribution patterns, and dynamic changes, and interpreting the underlying pathological implications reflected by these external signs. **Subsequently**, by integrating lesion information from multiple anatomical sites and time points, clinicians synthesize an overall assessment of the patient's dermatologic condition. This assessment is further combined with TCM dermatologic theories to systematically analyze disease etiology and pathogenesis, thereby capturing the essential characteristics and progression trends of the dermatologic disease. **On this basis**, physicians further incorporate the patient's medical history, constitutional features, and concurrent symptoms to comprehensively evaluate disease stage and overall severity, complete syndrome differentiation, and formulate corresponding therapeutic principles and TCM prescription strategies. **Consequently**, the diagnosis and treatment of dermatologic diseases in TCM can be viewed as a continuous clinical reasoning process that progresses from localized observation to holistic understanding, from surface manifestations to pathogenesis analysis, and from pathogenesis recognition to therapeutic decision-making.

Based on these clinical characteristics, the dermatologic diagnostic and therapeutic process in TCM can be restructured into three hierarchical levels of clinical cognition and decision-making:

1. **Primary Recognition of Localized Lesions**

Physicians establish an initial assessment of localized lesion states by observing lesion morphology, color, texture, and distribution, thereby providing foundational information for subsequent analysis. Accordingly, this level emphasizes the accurate recognition and semantic description of lesion-specific features derived from individual dermatologic presentations. **Therefore**, in AI modeling, the model is required to be capable of recognizing, understanding, and semantically expressing the localized lesion manifestations reflected by a single dermatologic image. Specifically, the model should extract clinically meaningful lesion information from visual features and generate corresponding textual descriptions, so as to demonstrate its preliminary cognitive ability regarding single-lesion morphology and related clinical phenomena.

2. **Global Dermatologic Interpretation and Specialty Pathogenesis Analysis**

Building upon localized recognition, physicians integrate dermatologic manifestations from multiple sites and perspectives to form a comprehensive understanding of the patient's overall lesion status. This process further involves applying TCM dermatologic theories to analyze the underlying etiology and pathogenesis reflected by the observed lesion patterns, thereby revealing the intrinsic pathological basis of the disease. **Accordingly**, based on primary recognition, the model is required to form a systematic understanding of the overall dermatologic status reflected by multi-view lesion images of the same patient. This process not only requires the model to aggregate and abstract information from multiple lesion images, but also to perform reasoning analysis of the corresponding specialty etiology and pathogenesis based on TCM dermatologic theory, and to demonstrate its comprehensive understanding of the patient's dermatologic condition by generating both global dermatologic specialty descriptions and pathogenesis analysis results.

3. **Comprehensive Diagnostic and Therapeutic Decision-Making**

With a clarified understanding of overall etiology and pathogenesis, physicians integrate medical history, physical signs, and current symptoms to determine disease stage and severity, finalize syndrome differentiation, and formulate corresponding therapeutic principles and prescription plans. **For AI models**, based on an established understanding of the patient's overall dermatologic status and specialty pathogenesis, the model should further incorporates clinical information such as medical history, physical signs, and current symptoms to comprehensively analyze and reason about the patient's current overall etiology and disease stage, and thereby generate a complete diagnostic and treatment plan, including syndrome differentiation, treatment principle selection, and prescription formulation.

Through aforementioned decomposition and redefinition, the TCM-based dermatologic diagnostic and treatment process is systematically articulated as a hierarchical, logically coherent clinical decision-making framework centered on holistic syndrome differentiation. This formulation establishes a clear clinical foundation for the standardized representation and methodological investigation presented in subsequent sections.

## 2.2 DERM-3R

We propose **DERM-3R** (**D**ermatology **E**xtraction and **R**easoning via **M**ulti-Agent), a multimodal multi-agent framework designed to achieve fine-grained lesion **r**ecognition, comprehensive dermatologic specialty **r**epresentation, and structured clinical **r**easoning for dermatologic diagnosis and treatment. The framework consists of three specialized agents: a **Dermatologic Recognition Agent (DERM-Rec)**, a **Dermatologic Representation Agent (DERM-Rep)**, and a **Dermatologic Clinical Reasoning Agent (DERM-Reason)**. All agents are built upon **Qwen2.5-VL-7B**, with its original language model replaced by **Tianyi**. It ensures that the resulting architecture incorporates the systematic TCM knowledge, including dermatologic theories and clinical application principles and practical experience. The modified model is referred to as **Derm-VL-7B**. All agents are trained using **LoRA**-based parameter-efficient fine-tuning. As illustrated in the left panel of Figure 1, the framework comprises the following three agents:

**Dermatologic Recognition Agent (DERM-Rec).** To enable effective perception of dermatologic conditions, the task is first decomposed into fine-grained recognition and semantic understanding of localized lesion morphology. At this stage, the agent receives a single dermatologic image as input and learns to extract clinically meaningful lesion features from localized visual patterns, thereby establishing foundational lesion-level cognition.

Specifically, all available lesion images from clinical psoriasis cases were collected, and dermatology specialists were invited to annotate each image $img_i$ with fine-grained corresponding localized lesion description, $d_i$, based on standard clinical observation dimensions. Following systematic verification and quality control, each lesion image was paired with its textual description, resulting in a dataset comprising **518 single-image–text pairs,** $\boldsymbol{D_{Rec}} = \{img_i, d_i\}_{i=0}^{518}$ . Using **Derm-VL-7B** as the base model, the maximum likelihood objective for training DERM-Rec to learn to generate localized lesion descriptions from a single lesion image, as shown in Eq. (1) and (2). The resulting agent is defined as the **Dermatologic Recognition Agent (DERM-Rec)**.

$$p_{\theta_{Rec}}(d_i|img_i) = \prod_j^T p_{\theta_{Rec}}(d_i^j|d_i^{j-1}, img_i) \tag{1}$$

$$\mathcal{L}_{\theta_{Rec}} = -\mathbb{E}_{(img_i,d_i)\sim D_{Rec}}[\sum_{j=1}^{|D_{Rec}|} logp_{\theta_{Rec}}(d_i|img_i)] \tag{2}$$

**Dermatologic Representation Agent (DERM-Rep)**. Building upon DERM-Rec, the modeling objective is extended from localized lesion recognition to comprehensive interpretation and representation of multi-view dermatologic information. In real-world clinical settings, a patient encounter typically includes multiple lesion images acquired from different anatomical sites or perspectives, rendering single-image analysis insufficient for capturing the patient's overall dermatologic condition.

Accordingly, all lesion images collected during a single clinical visit are treated as a unified input set. Dermatology specialists then generate detailed, patient-level dermatologic descriptions based on these image sets. Since surface manifestations alone do not fully capture disease essence, expert clinicians further provide corresponding TCM dermatologic pathogenesis analyses to explain the underlying mechanisms reflected by the observed lesion patterns. This process yields a kind of reasoning samples with multi-image $\boldsymbol{x}$ inputs and dual outputs consisting of global lesion descriptions $\widehat{d}_i$ and pathogenesis analyses $m$. Ultimately, **148 high-quality samples** were constructed, denoted as $D_{Rep}$, to train the agent to aggregate, abstract, and semantically model multi-view dermatologic information. This agent inherits single-lesion recognition capabilities while forming structured representations of patient-level dermatologic states and corresponding TCM pathogenesis. We refer to this agent as the **Dermatologic Representation Agent (DERM-Rep)**. To explicitly model the two-step reasoning process underlying dermatologic interpretation, DERM-Rep factorizes the conditional distribution as:

$$p_{\theta_{Rep}}(\widehat{d}_i, m|\boldsymbol{x}) = p_{\theta_{Rep}}(\widehat{d}_i|\boldsymbol{x}) \cdot p_{\theta_{Rep}}(m|\boldsymbol{x}, \widehat{d}_i) \tag{3}$$

The corresponding training objective is defined as:

$$\mathcal{L}_{\theta_{Rep}} = -\mathbb{E}_{(\boldsymbol{x},\widehat{d}_i,m)\sim D_{Rep}}[logp_{\theta_{Rep}}(d_i|\boldsymbol{x}) + logp_{\theta_{Rep}}(m|\boldsymbol{x}, d_i)] \tag{4}$$

**Dermatologic Clinical Reasoning Agent (DERM-Reason)**. Through the preceding stages, DERM-Rec and DERM-Rep establish lesion-level perception and patient-level representation of dermatologic conditions and pathogenesis. On this basis, a multimodal clinical reasoning agent is constructed to address high-level diagnostic and therapeutic decision-making in real-world clinical scenarios.

Specifically, the patient-level dermatologic descriptions $\widehat{d}_i$ and TCM pathogenesis analysis $m$ produced by DERM-Rep are treated as intermediate reasoning representations and integrated with multi-view lesion images $\boldsymbol{x}$, medical history $h$, physical signs $idx$, and current symptoms $s$ to form structured multimodal pathological inputs. The outputs consist of **overall TCM pathogenesis analysis** $M$, **syndrome differentiation** $S$, **treatment principles** $T$, and **TCM prescription** $P$. It forms a dataset $D_{reason}$ comprising 134 verified samples. A two-stage reasoning-based generation strategy is adopted. The agent first produces an overall pathogenesis analysis as an intermediate high-level clinical representation, which subsequently constrains and guides the generation of syndrome diagnosis, treatment principles, and prescriptions. This design mirrors the clinical reasoning pathway from pathogenesis analysis to therapeutic decision-making. The resulting agent is termed the **Dermatologic Clinical Reasoning Agent (DERM-Reason)**. The hierarchical nature of clinical reasoning progress we design to model the generation process is as a two-level conditional decomposition:

$$p_{\theta_{reason}}(M,S,T,P|\boldsymbol{x},h,idx,s,\widehat{d}_i,m) = p_{\theta_{reason}}(M|u) \cdot p_{\theta_{reason}}(S,T,P|u,M) \quad (5)$$

Where $u = (\boldsymbol{x},h,idx,s,\widehat{d}_i,m)$. The training objective is defined as:

$$\mathcal{L}_{\theta_{Rep}} = -\mathbb{E}_{(u,M,S,T,P)\sim D_{reason}}[logp_{\theta_{reason}}(M|u) + logp_{\theta_{reason}}(S,T,P|u,M)] \quad (6)$$

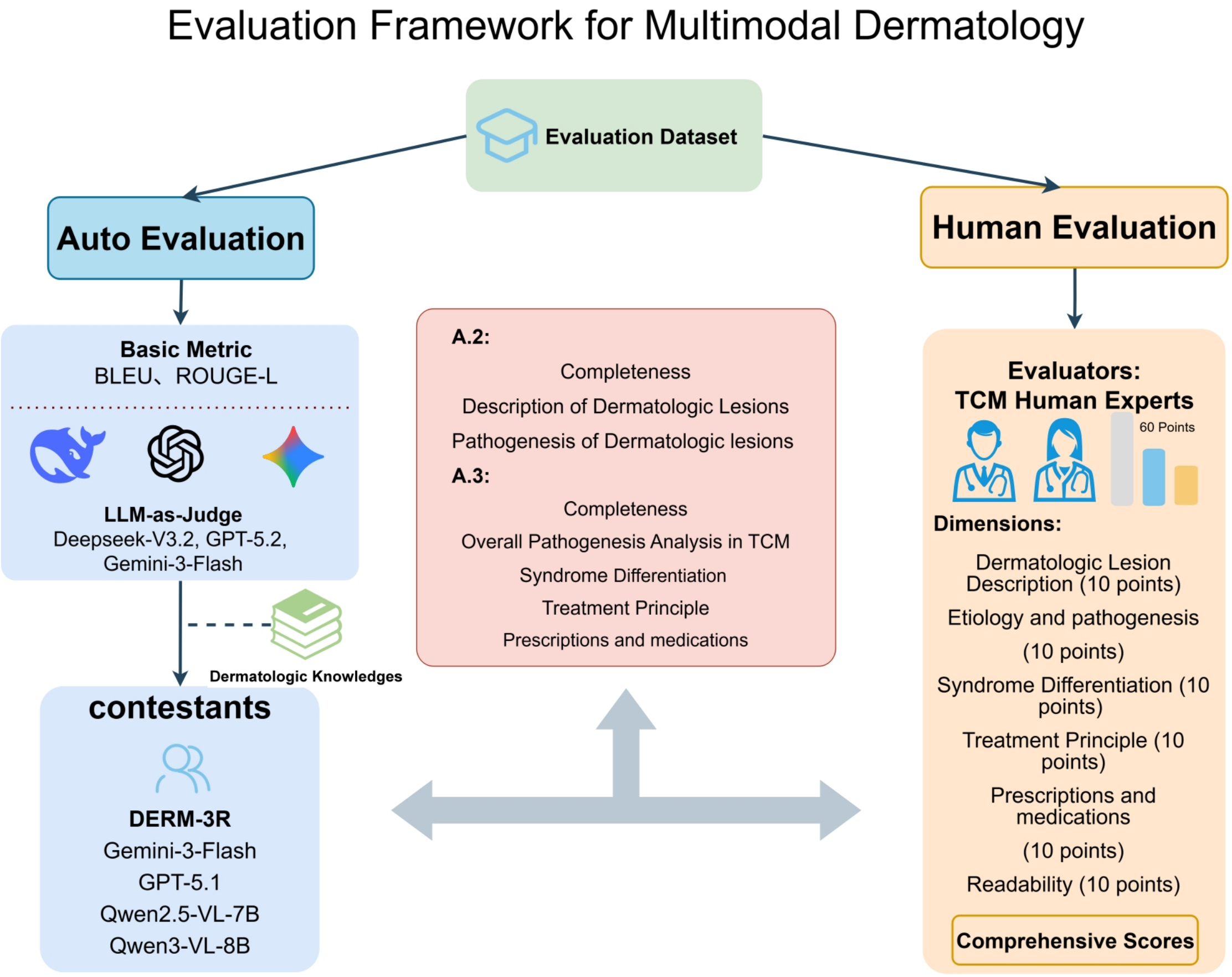

Fig. 2. The evaluation framework for the multimodal multi-agent in this work. The DERM-Rep and DERM-Reason are designed for the in-demand to solve the challenges in real-world clinical challenges. Thus, we main evaluate the performances of agents DERM-Rep and DERM-Reason. The evaluation framework consists of two parts: the automatic evaluation and Human doctor evaluations. The automatic evaluation contains the basic metrics: BLEU-4 and ROUGE-L, and the LLM-as-a-Judge strategy that enhanced with the dermatologic knowledge base for the accurate expertise in general-purpose LLMs. The human doctor evaluation is designed as a multi-center cross-validation evaluation method involving **9** hospitals. We enroll **1-2** board-certified dermatologists with senior or associate senior professional titles to evaluate the task-wised and total scores of each model's responses on each test samples. Then, the average scores are computed as the final scores. The automated

evaluation provides objective, bias-free results, while human clinicians offer comprehensive and thoughtful assessments of all comparison models under conditions designed to minimize subjective bias from a single doctor.

### Evaluation Framework

To comprehensively assess the performance of the proposed DERM-3R framework in multimodal dermatologic diagnosis and treatment, we designed a hybrid evaluation protocol that integrates automatic metrics and medical expert-level assessments, taking into full consideration the complexity of clinical dermatologic reasoning and the requirements for objective evaluation. As the primary focus of this study lies in clinical diagnosis and therapeutic decision-making, we conducted extensive evaluations on the **DERM-Rep** and **DERM-Reason** agents. The DERM-Rec agent, which addresses a relatively preliminary task of single-image lesion recognition and understanding, was therefore excluded from the evaluation. For the concrete results from the view of real TCM dermatologic case, we demonstrate the cases collected from the Gulou Hospital of Traditional Chinese Medicine of Beijing and generate the all results by DERM-3R, which can be found in **Appendix B**.

### Automatic Evaluation

The objective of automatic evaluation is to quantitatively assess the similarity between model-generated outputs and gold-standard annotations by using fixed, non-interactive metrics, thereby providing an unbiased estimation of model performance. Both **DERM-Rep** and **DERM-Reason** involve multiple natural language generation (NLG) subtasks. Accordingly, we **first** employed widely adopted lexical overlap metrics, including **BLEU** and **ROUGE-L**, to evaluate the overall quality of generated outputs. **In addition**, we incorporated an **LLM-as-a-Judge** paradigm to enable more fine-grained and semantically informed automatic evaluation. Notably, to mitigate the potential generalization limitations of **LLM-as-a-Judge** when evaluating **out-of-distribution (OOD)** domain-specific tasks[21,22], enhancing its professional rigor of evaluations conducted on specialist AI models, we constructed a **domain-specific knowledge base for TCM dermatology**. This knowledge base was integrated into the judging process via a **retrieval-augmented generation (RAG)** mechanism, enabling the judge models to reference authoritative TCM dermatologic knowledge prior to scoring. When combined with gold-standard annotations, this approach substantially improves the professional relevance, reliability, and interpretability of the automatic evaluation results.

**Evaluation of DERM-Rep.** Given that DERM-Rep operates on multimodal inputs and outputs, we selected general-purpose LLMs capable of jointly processing both images and text as judge models. Specifically, **Gemini-3-Flash** and **GPT-5.2** were employed to evaluate the outputs of DERM-Rep. The evaluation focused on the completeness of the generated responses, the accuracy of dermatologic lesion descriptions, and the correctness of the corresponding TCM pathogenesis analysis. The evaluation dimensions for DERM-Rep include:

1. **Accuracy of dermatologic lesion descriptions**;
2. **Accuracy of the generated TCM dermatologic pathogenesis analysis**.

The detailed scoring rubric for DERM-Rep is provided in **Appendix A.1**.

**Evaluation of DERM-Reason.** Based upon the judge models used for DERM-Rep, we further incorporated DeepSeek-V3.2 as an additional judge model to diversify the evaluation ensemble and reduce potential model-specific biases. The evaluation of DERM-Reason focuses on five critical aspects of clinical reasoning and decision-making:

1. **Overall pathogenesis analysis**: assessing whether the model effectively integrates visual

features (e.g., erythema, pigmentation) with other clinical information to produce a coherent and clinically meaningful pathogenesis reasoning process;

2. **Syndrome differentiation**: evaluating the correctness of the generated syndrome diagnosis;
3. **Treatment principle selection**: assessing whether the selected treatment principles are appropriate, syndrome-consistent, and sufficiently cover the inferred pathogenesis;
4. **Formula selection and prescription generation**: evaluating the accuracy of formula matching and the correctness of herbal composition, computed according to the proposed quantitative formulation-matching criteria;
5. **Output completeness**: assessing whether all required diagnostic and therapeutic components are present.

The detailed evaluation protocol and scoring criteria for DERM-Reason are presented in **Appendix A.2**.

### Human Evaluation

Due to automated evaluations are inherently limited by the professional knowledge encoded within the underlying models or the fixed NLG metrics, they cannot fully capture the quality of long, semantically rich, and interdependent outputs required for complex clinical tasks. To obtain a more comprehensive and expert-informed assessment, we designed a **multicenter cross-validation evaluation strategy** involving nine hospitals. Each participating institution invited one to two dermatology clinicians with extensive clinical experience and intermediate or senior professional titles. Following the evaluation criteria we provided, these clinicians independently assessed all participating models and generated both qualitative and quantitative evaluations based on their own clinical expertise and diagnostic – therapeutic experience. The purpose of this multicenter assessment strategy is to obtain a broad, clinically grounded evaluation from domain experts while reducing individual evaluator bias. In total, **15 dermatology clinicians** participated in this multicenter cross-validation evaluation. All clinicians are provided the **blind model name with A to E** for all comparison models. The evaluation criteria include: **Dermatologic Lesion Description** (10 points), **Overall Etiology and pathogenesis** (10 points), **Syndrome Differentiation** (10 points), **Treatment Principle** (10 points), **Prescriptions and medications** (10 points), **Readability** (10 points), with a total score of 60 points[23].

### Comparison Models

In this study, to align with the characteristics of real-world dermatologic clinical practice, we selected **LLMs** capable of jointly processing visual and textual information as baseline models for comparative evaluation. Accordingly, we chose four representative general-purpose multimodal LLMs: **Gemini-3-Flash**, **GPT-5.1-instant**, **Qwen2.5-VL-7B**, and **Qwen3-VL-8B**.

Among these baselines, **Gemini-3-Flash** and **GPT-5.1** represent **large-scale, hundred-billion-parameter general-purpose models**, while **Qwen2.5-VL-7B** and **Qwen3-VL-8B** are models of comparable scale to the model fine-tuned with DERM-3R framework. Qwen2.5-VL-7B was selected because it serves as the **base model** for DERM-3R and thus provides a necessary reference point for performance comparison. In contrast, **Qwen3-VL-8B** was included for comparison due to the consideration of performance comparison of the latest generation of the Qwen series on a dermatology-specific task, since the Qwen3 family had not yet been released at the time our work was studied. We did not include larger-parameter variants of the Qwen3 series in our evaluation due to the other two representative large-scale multi-modal models had been included. This decision was also decided by practical considerations regarding **limited computational resources in clinical settings**, where the deployment and training of substantially larger models may not be feasible. Consequently, comparisons with such large-scale models were considered less representative of real-world clinical applicability and were therefore excluded from this study.

## Experiments

### Automatic Evaluation Using BLEU-4 and ROUGE-L

Table 1. Evaluations for integrated dermatologic lesion description and TCM pathogenesis analysis. The highest score of each item is in bold.

| Evaluations for Description and Pathogenesis of Dermatologic Lesions | | | | | | |
|---|---|---|---|---|---|---|
| | **Item** | **GPT5.1-instant** | **Gemini-3-Flash** | **Qwen2.5-VL-7B** | **qwen3-VL-8B** | **DERM-Rep** |
| **BLEU-4** | Description of Dermatologic Lesions | 0.0714 | 0.0584 | 0.0543 | 0.0354 | **0.2298** |
| | Pathogenesis of Dermatologic Lesions | 0.0105 | 0.0102 | 0.0658 | 0.0243 | **0.1246** |
| | Total | 0.0410 | 0.0343 | 0.0600 | 0.0298 | **0.1772** |
| **Rouge-L** | Description of Dermatologic Lesions | 0.2331 | 0.2324 | 0.3325 | 0.2662 | **0.4786** |
| | Pathogenesis of Dermatologic Lesions | 0.1758 | 0.1423 | 0.2682 | 0.1767 | **0.3763** |
| | Total | 0.2044 | 0.1874 | 0.3004 | 0.2214 | **0.4275** |

Table 1 reports the BLEU-4 and ROUGE-L scores for DERM-Rep and all baseline models on the two subtasks of **integrated dermatologic lesion description** and **TCM pathogenesis analysis**. DERM-Rep consistently achieves the highest scores across all metrics. For lesion description, DERM-Rep obtains a BLEU-4 score of 0.2298 and a ROUGE-L score of 0.4786, substantially outperforming GPT-5.1 (BLEU-4: 0.0714; ROUGE-L: 0.2044) and Gemini-3-Flash (BLEU-4: 0.0584; ROUGE-L: 0.1874). Compared with same-scale multimodal models, DERM-Rep exceeds Qwen2.5-VL-7B (BLEU-4: 0.0543; ROUGE-L: 0.1987) and Qwen3-VL-8B (BLEU-4: 0.0354; ROUGE-L: 0.1736) by a large margin. A similar trend is observed for TCM pathogenesis analysis. DERM-Rep achieves a **BLEU-4 score of 0.1246** and a **ROUGE-L score of 0.3763**, whereas all baseline models remain below 0.07 in BLEU-4 and below 0.25 in ROUGE-L. When averaging across both subtasks, DERM-Rep reaches an overall **BLEU-4 score of 0.1772** and **ROUGE-L score of 0.4275**, approximately **2–5× higher** than general-purpose multimodal LLMs. These results indicate that DERM-Rep not only improves surface-level textual similarity, but also captures domain-specific structure and terminology required for dermatologic lesion aggregation and pathogenesis reasoning.

Table 2. the BLEU-4 and ROUGE-L results for DERM-Reason across five clinical subtasks: overall pathogenesis analysis of a patient, syndrome differentiation, treatment principle selection, formula selection, and prescription generation. DERM-Reason achieves the highest overall performance among all evaluated models. The highest score of each item is in bold.

| Evaluations for Final Diagnosis and Treatment Strategy | | | | | | |
|---|---|---|---|---|---|---|
| | **Item** | **GPT5.1-Instant** | **Gemini-3-Flash** | **Qwen2.5-VL-7B** | **qwen3-VL-8B** | **DERM-Reason** |
| **BLEU-4** | Overall pathogenesis analysis | 0.1293 | 0.0379 | 0.1299 | 0.1615 | **0.1930** |
| | Syndrome differentiation | 0.0884 | 0.0589 | 0.1070 | 0.0868 | **0.1254** |
| | Treatment principle selection | 0.1747 | 0.0729 | 0.2016 | 0.2075 | **0.4877** |
| | Formula selection | 0.0487 | 0.0574 | 0.0518 | 0.0363 | **0.3997** |
| | TCM prescription generation | 0.3908 | 0.4340 | 0.3325 | **0.4381** | 0.2379 |
| | Average | 0.1664 | 0.1322 | 0.1646 | 0.1861 | **0.2887** |
| **Rouge-L** | Overall pathogenesis analysis | 0.3725 | 0.2276 | 0.3323 | 0.4066 | **0.4134** |
| | Syndrome differentiation | 0.5036 | 0.3879 | 0.4714 | 0.4433 | **0.6111** |
| | Treatment principle selection | 0.4674 | 0.3847 | 0.5164 | 0.5360 | **0.7424** |
| | Formula selection | 0.2600 | 0.2105 | 0.3359 | 0.2551 | **0.7778** |
| | TCM prescription generation | 0.3092 | 0.3135 | 0.3042 | **0.3603** | 0.3562 |
| | Average | 0.3735 | 0.3048 | 0.3921 | 0.2809 | **0.5802** |

We collect the BLEU-4 and ROUGE-L results for DERM-Reason across five clinical subtasks in Table 2: *overall pathogenesis analysis*, *syndrome differentiation*, *treatment principle selection*, *formula selection*, and *TCM prescription generation*. DERM-Reason achieves the highest overall performance among all evaluated models. Across all subtasks, DERM-Reason attains an average BLEU-4 score of 0.2887 and an average ROUGE-L score of 0.5802, outperforming GPT-5.1 (BLEU-4: 0.2164; ROUGE-L: 0.3735), Gemini-3-Flash (BLEU-4: 0.1987; ROUGE-L: 0.3521), and same-scale Qwen models. For *overall pathogenesis analysis*, DERM-Reason achieves a **BLEU-4 score of 0.1930** and a **ROUGE-L score of 0.4134**, markedly higher than GPT-5.1 (BLEU-4: 0.1293; ROUGE-L: 0.3725) and Qwen2.5-VL-7B (BLEU-4: 0.1299; ROUGE-L: 0.3323). This indicates that DERM-Reason more accurately reproduces the structured causal explanations present in expert annotations. In *syndrome differentiation*, DERM-Reason reaches a **BLEU-4 score of 0.1254** and a **ROUGE-L score of 0.6111**, outperforming all baselines by a clear margin. Baseline models frequently produce plausible but overly generic syndrome labels, resulting in lower lexical and structural overlap with gold annotations. It's also worthy to notice that, for ***overall pathogenesis analysis*** and **Syndrome Differentiation**, Qwen series multimodal models with smaller model scale achieve the higher BLEU-4 and ROUGE-L scores than GPT-5.1 and Gemini-3-Flash. It demonstrates better consistent content generation of Qwen series models along with clinical descriptions. For *Treatment Principle Selection*, DERM-Reason demonstrates particularly strong performance in treatment principle selection, achieving a BLEU-4 score of 0.4877 and a ROUGE-L score of 0.7424. In contrast, GPT-5.1 records BLEU-4 and ROUGE-L scores of 0.2113 and 0.4028, respectively. This substantial gap suggests that conditioning treatment reasoning on explicitly generated pathogenesis representations improves logical consistency. For *formula selection*, DERM-Reason achieves a BLEU-4 score of 0.3997 and a ROUGE-L score of 0.7778, significantly exceeding all baseline models. For *TCM prescription generation*, however, DERM-Reason's BLEU-4 score (0.2379) is closer to that of some baseline models, while its ROUGE-L score (0.3562) remains competitive. This discrepancy reflects the inherent limitations of n-gram-based metrics in evaluating highly open-ended prescription texts, where multiple valid formulations may exist.

## Automatic Evaluation by LLM-as-a-Judge

We first analyze the LLM-as-a-Judge evaluation results for agent **DERM-Rep**, which is designed to generate integrated dermatologic lesion descriptions and corresponding TCM pathogenesis analyses from multi-view lesion images as shown in Fig. 3. It reports the judge-based scores under two independent evaluators, **Gemini-3-Flash** and **GPT-5.2**, across multiple evaluation dimensions. Under the **Gemini-3-Flash Judge**, DERM-Rep achieves a total score of **22.2500**, ranking closely behind GPT-5.1-Instant (**23.1667**) and Gemini-3-Flash (**22.8333**), while clearly outperforming the base model Qwen2.5-VL-7B (**18.1667**) and maintaining a competitive score relative to Qwen3-VL-8B (**21.8333**). A similar pattern is observed under the **GPT-5.2 Judge**, where DERM-Rep obtains a total score of **14.2500**, substantially exceeding Qwen2.5-VL-7B (**11.7500**) but remaining slightly lower than GPT-5.1-Instant (**15.8750**) and Qwen3-VL-8B (**14.9167**). Averaged across judges, DERM-Rep attains a mean total score of **18.2500**, compared to **14.9583** for its base model, indicating a clear performance gain attributable to the DERM-3R framework.

A dimension-level analysis reveals that **DERM-Rep's advantage is most pronounced in TCM pathogenesis reasoning**. Under the GPT-5.2 Judge, DERM-Rep achieves the highest score in *Pathogenesis of the skin lesion* (**4.0833**), outperforming GPT-5.1-Instant (**3.9583**) and Gemini-3-Flash (**3.4583**). This suggests that the agent's structured aggregation of multi-view lesion information facilitates more clinically consistent and professionally aligned pathogenesis explanations, particularly when evaluated under stricter judging criteria. In contrast, DERM-Rep exhibits relatively lower scores in *Skin lesion characteristics* (e.g., **5.1667** under GPT-5.2), compared with larger general-purpose models such as GPT-5.1-Instant (**6.9167**) and Gemini-3-Flash (**6.5000**). This indicates that while DERM-Rep excels in higher-level explanatory reasoning, its fine-grained coverage of detailed lesion morphology may be less aligned with the lexical or descriptive preferences captured by some judge models. Such performance difference might be caused by the multi-modal semantic alignment capabilities between general-purpose hundred-billion-parameter general-purpose multimodal models and Qwen2.5-VL-7B model.

Overall, the LLM-as-a-Judge evaluation indicates that **DERM-Rep provides a robust improvement over same-scale multimodal baselines**, particularly in clinically meaningful pathogenesis reasoning, while leaving room for further refinement in fine-grained lesion description consistency.

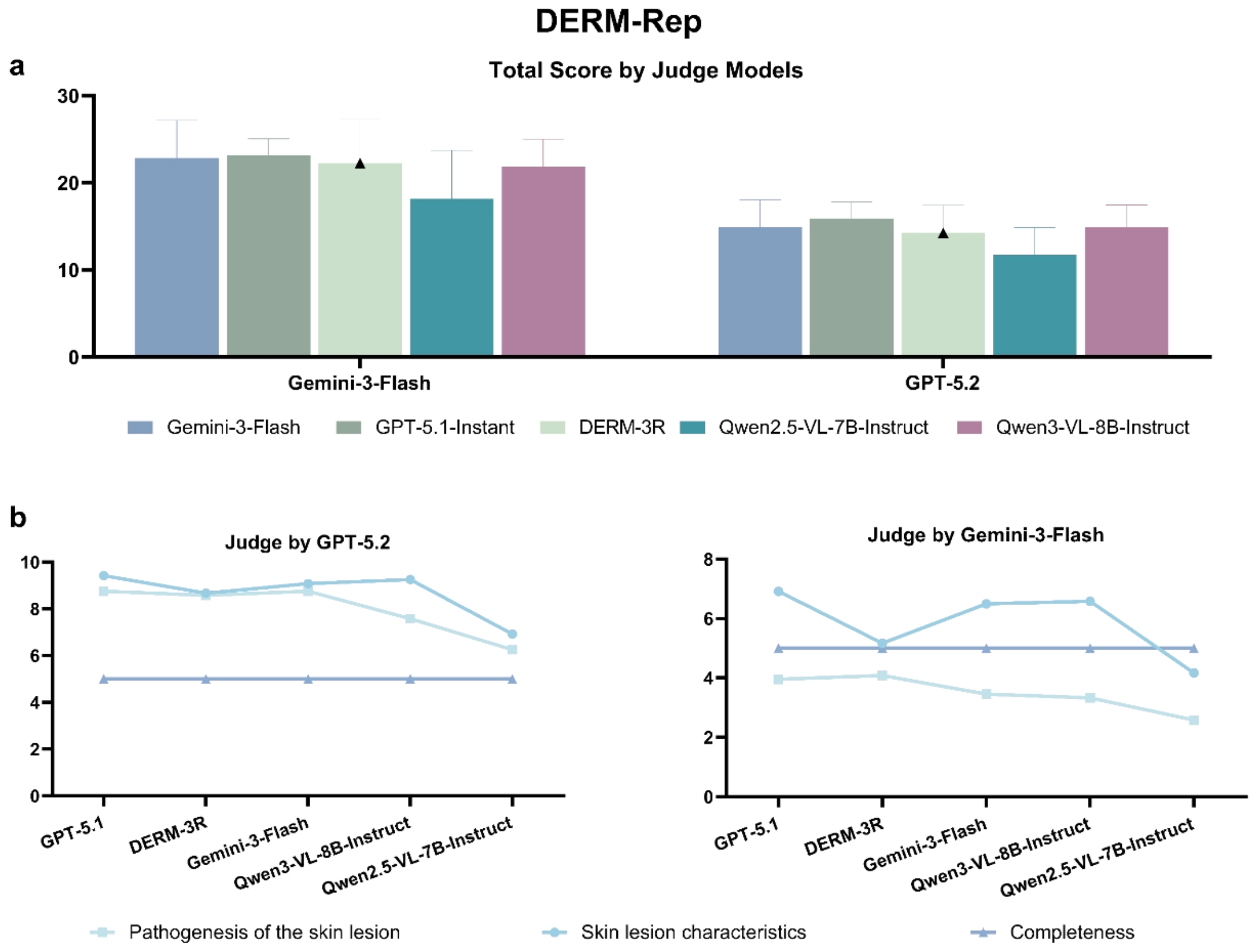

Fig. 3. The evaluation results for all comparisons with agent DERM-Rep. The total scores and item-based scores are illustrated in part a and b, respectively. As shown in part a, DERM-Rep achieves the comparable performances with GPT-5.1-instant and Gemini-3-Flash, and the better performance than the same parameter scale with Qwen series models such as Qwen2.5-VL-7B-instruct (the base model for DERM-3R) and Qwen3-VL-8B-instruct in total scores. As for the performances from detailed items, DERM-Rep shows its greatest advantage in TCM pathogenesis reasoning, achieving the highest pathogenesis score under the GPT-5.2 Judge and outperforming larger general-purpose models like GPT-5.1-Instant and Gemini-3-Flash. This suggests that its structured aggregation of multi-view lesion information leads to more clinically coherent pathogenesis explanations, though its performance on detailed skin lesion characteristics remains weaker and less aligned with the lexical/descriptive preferences of some judge models.

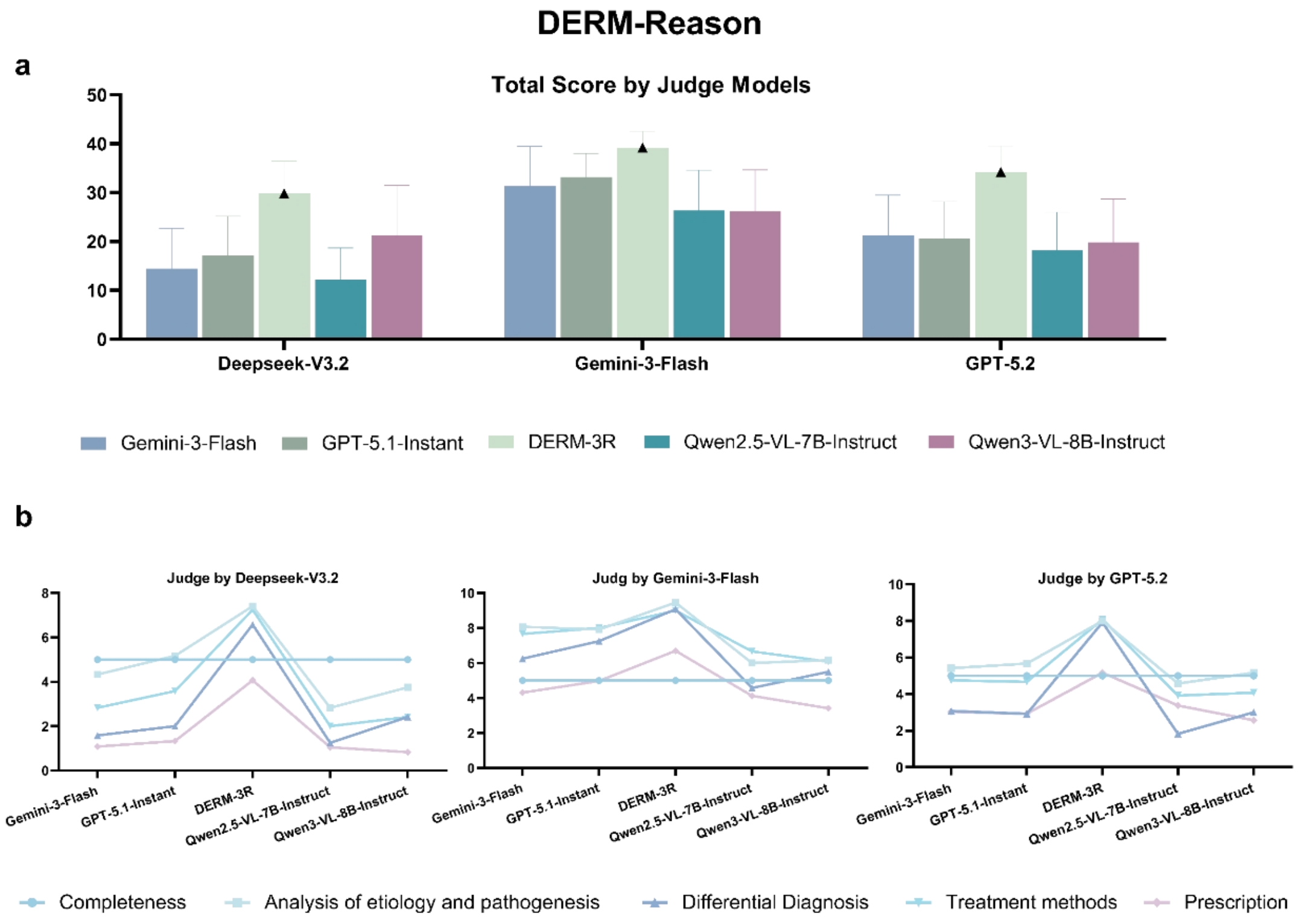

Fig. 4. The evaluation results for all comparisons with agent DERM-Reason. It presents the LLM-as-a-Judge evaluation results for DERM-Reason, which performs end-to-end multimodal clinical reasoning, including overall pathogenesis analysis, syndrome differentiation, treatment principle selection, and prescription recommendation. Evaluations were conducted independently using three judge models — Deepseek-V3.2, Gemini-3-Flash, and GPT-5.2 — to reduce evaluator-specific bias. Across all judges, DERM-Reason consistently achieves the highest total scores (29.8500, 39.2442, and 34.2158, respectively), substantially outperforming the second-best model, GPT-5.1-Instant, as well as all other baseline multimodal models. Averaged across judges, DERM-Reason attains a mean total score of 34.4366, compared with 23.6462 for GPT-5.1-Instant and 18.8866 for the base model Qwen2.5-VL-7B. Task-wise analysis further shows that DERM-Reason demonstrates systematic advantages across all core reasoning dimensions, achieving higher average scores in overall pathogenesis analysis (8.2917), syndrome differentiation (7.8611), treatment principle selection (8.1111), and prescription generation (5.3116) than all comparison models. The Completeness dimension is saturated across models (score = 5), indicating that performance differences primarily arise from content quality and reasoning coherence rather than output formatting. In addition, DERM-Reason exhibits lower score variance across cases (average standard deviation ≈ 5) than GPT-5.1-Instant (≈ 7), suggesting improved robustness and reduced sensitivity to case-level variability.

We next evaluate agent **DERM-Reason**, which performs end-to-end multimodal clinical reasoning, including overall pathogenesis analysis, syndrome differentiation, treatment principle selection, and prescription recommendation. Table 3 summarizes the LLM-as-a-Judge results under three independent judges: **Deepseek-V3.2**, **Gemini-3-Flash**, and **GPT-5.2**. Across all judges, DERM-Reason consistently achieves the highest total scores among all evaluated models. Specifically, DERM-Reason attains total scores of **29.8500**, **39.2442**, and **34.2158** under Deepseek-V3.2, Gemini-3-Flash, and GPT-5.2, respectively. These performances substantially exceed those of GPT-5.1-Instant (**17.1111**, **33.1433**, **20.6856**) which achieves the second-place performance and all other baselines. Averaged across judges, DERM-Reason achieves a mean total score of **34.4366**, compared with **23.6462** for GPT-5.1-Instant and **18.8866** for the base model Qwen2.5-VL-7B, demonstrating a clear and stable performance advantage.

Task-wise analysis further highlights the systematic strengths of DERM-Reason. In **overall pathogenesis analysis**, DERM-Reason achieves an average score of **8.2917** across three judges, outperforming GPT-5.1-Instant (**6.2500**) and Gemini-3-Flash (**5.9444**), indicating more coherent and causally structured explanations. In **syndrome differentiation**, DERM-Reason attains a mean score of **7.8611**, nearly doubling the score of GPT-5.1-Instant (**4.0556**), reflecting improved alignment with expert diagnostic reasoning. Similarly, in **treatment principle selection**, DERM-Reason achieves an average score of **8.1111**, compared to **5.4167** for GPT-5.1-Instant, suggesting stronger logical consistency between inferred pathogenesis and therapeutic decisions. For **prescription generation**, a task characterized by high openness and multiple valid solutions, DERM-Reason still maintains a clear advantage, with an average score of **5.3116**, compared with **3.0814** for GPT-5.1-Instant. This indicates that the model not only generates plausible prescriptions but does so in a manner judged to be more clinically coherent and appropriate. The *Completeness* dimension is saturated across models (score = 5), suggesting that performance differences primarily stem from content quality rather than structural formatting. In terms of robustness, DERM-Reason also demonstrates relatively stable performance, with an average standard deviation of approximately **5.08**, lower than that of GPT-5.1-Instant (**~6.80**), indicating reduced sensitivity to case-level variability.

Taken together, the LLM-as-a-Judge evaluation provides strong evidence that **DERM-Reason substantially outperforms both general-purpose and same-scale multimodal models across all major clinical reasoning dimensions**. The consistent superiority across multiple independent judges underscores the effectiveness of the DERM-3R multi-agent framework in structuring complex dermatologic reasoning under limited data and computational resources.

Table 3. The variances of all comparison models in all six tasks in human evaluation. The DERM-3R achieves the lowest variances among all tasks except **Dermatologic Lesion Description** and **Analysis of Etiology and Pathogenesis in TCM** tasks. The lowest variance for each task is in bold.

| | Dermatologic Lesion Description | Analysis of Etiology and Pathogenesis in TCM | Syndrome Differentiation | Treatment Principle Selection | Prescriptions and Medications | Readability |
|---|---|---|---|---|---|---|
| **Qwen3-VL-8B-Instruct** | 1.75 | 1.93 | 2.00 | 2.02 | 1.87 | 1.55 |
| **Qwen2.5-VL-7B-Instruct** | 2.00 | 2.15 | 2.08 | 2.09 | 1.95 | 1.65 |
| **Derm-3R** | 1.62 | 1.56 | **1.53** | **1.54** | **1.40** | **1.14** |
| **GPT-5.1-Instant** | **1.54** | **1.45** | 1.57 | 1.59 | 1.60 | 1.35 |
| **Gemini-3-Flash** | 1.81 | 1.58 | 1.83 | 1.82 | 1.70 | 1.41 |

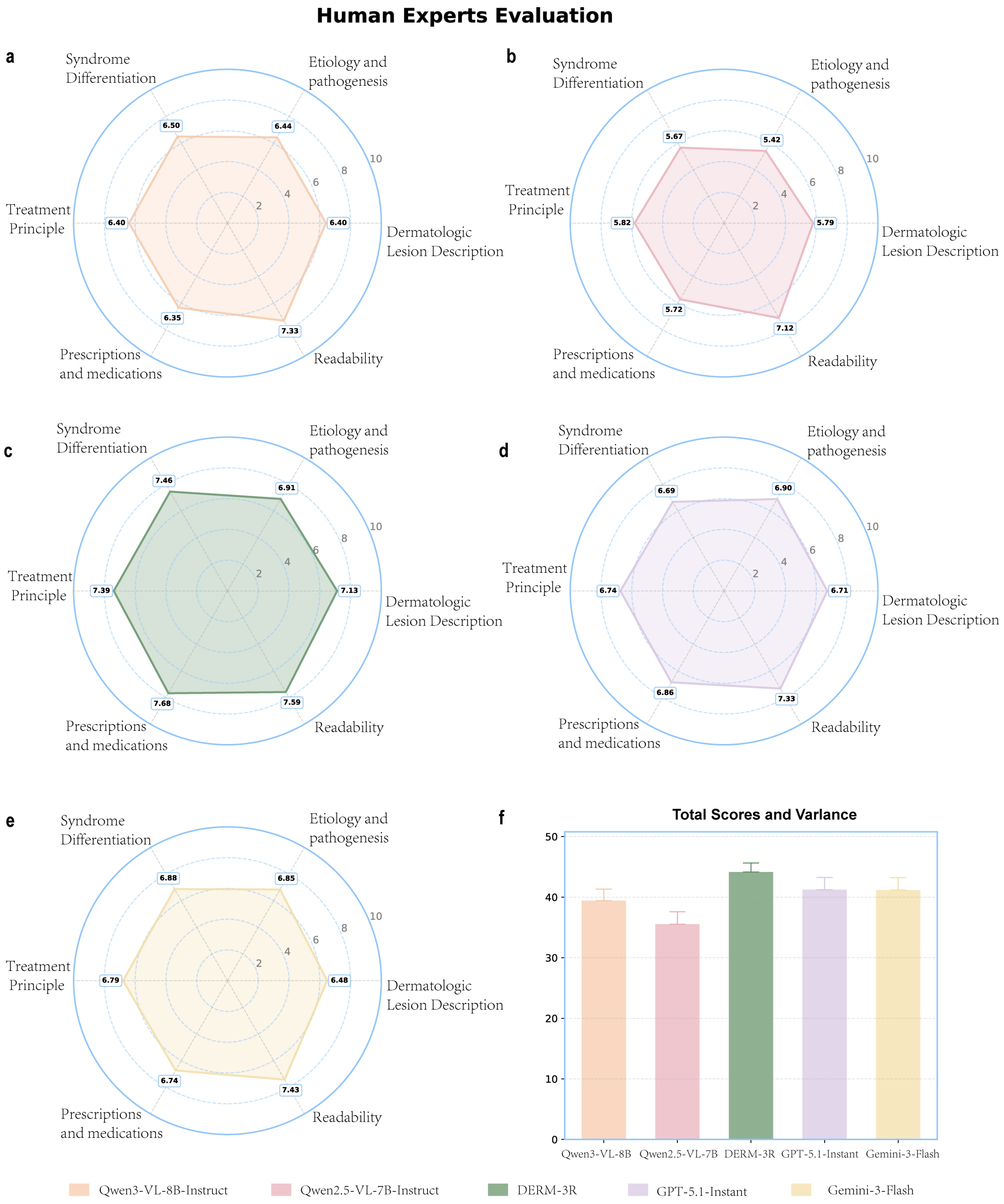

Fig. 5. Multicenter human evaluation of DERM-3R and baseline models. It presents the results of a multicenter human evaluation in which outputs from DERM-Rep and DERM-Reason were combined into a single comprehensive report per clinical case and assessed by 15 dermatology clinicians from 9 hospitals. Panels a–e show mean scores for six evaluation dimensions—dermatologic lesion description, over analysis of etiology and pathogenesis in TCM, syndrome differentiation, treatment principle selection, prescriptions and medications, and readability—while panel f summarizes the total scores. Overall, DERM-3R achieves the highest total score (44.16) with the lowest variance (1.49), outperforming Qwen3-VL-8B-instruct (39.42, var 1.91), Qwen2.5-VL-7B-instruct (35.54, var 2.06), GPT-5.1-Instant (41.23, var 2.03), and Gemini-3-Flash (41.17, var 2.05), indicating both superior performance and greater stability. Across all five dimensions except readability, DERM-3R attains the highest average scores with consistently low variances, with particularly notable gains in dermatologic lesion description, which is critical for bridging modern dermatologic terminology with TCM etiology–pathogenesis reasoning and thereby enables downstream improvements in TCM analysis, syndrome differentiation, treatment selection, and prescription generation. Compared with its base model Qwen2.5-VL-7B, DERM-3R achieves larger performance gains across all tasks than Qwen3-VL-8B, reflecting the effectiveness of the DERM-3R training strategy in enhancing lesion expression and downstream decision-making. Qwen3-VL-8B-instruct, although slightly below GPT-5.1-Instant and Gemini-3-Flash in total score, performs comparably to these general-purpose models, likely due to its specialization for TCM multimodal scenarios, but still exhibits relatively high variance. GPT-5.1-Instant and Gemini-3-Flash display task-dependent advantages—GPT-5.1-Instant performing better in TCM etiology/pathogenesis analysis and prescriptions, and Gemini-3-Flash in syndrome differentiation—while GPT-5.1-Instant shows consistently lower variance than Gemini-3-Flash, suggesting more robust and stable outputs.

## Human Evaluation

For human evaluation, we combine the results of DERM-Rep with DERM-Reason, forming a comprehensive results for each clinical case. The results of human evaluation are demonstrated in Fig. 5. Their variances among 15 clinicians are shown in Table 3. As shown in part **f**, we can notice that DERM-3R obtains the absolute performance margin than other all comparison models. It achieves 44.16 total scores with 1.49 variance, while Qwen3-VL-8B-instruct, Qwen2.5-VL-7B-instruct, GPT-5.1-instant, and Gemini-3-flash obtains the total scores and variances of 39.42 (1.91), 35.54 (2.06), 41.23 (2.03), and 41.17 (2.05), respectively. DERM-3R obtains the highest total score with the lowest variance, which shows its substantial performance and the consistent stability. Compared Qwen3-VL-8B with Qwen2.5-VL-8B, the total score of Qwen3-VL-8B is improved with 10.1% (3.88 absolute score) in special domain tasks, indicating the increased in-domain problem-solving ability. However, such problem-solving ability and the stability still limited due to the variance is decreased with 7.4% not decreased significantly. While the DERM-3R improve the performance with 24.3% (8.63 absolute score) while the variance is decreased with 27.5%, demonstrating the efficacy and effectiveness of DERM-3R. Besides, Qwen3-VL-8B can generate more stable and suitable content according to the TCM case and dermatologic images than GPT-5.1-Instant and Gemini-3-Flash even the total score is less than them. It suggests that researchers should select different general-purpose model as their base model in a concrete medical scenario.

For the concrete items as shown from parts **a** to **e**, DERM-3R performs consistently the highest average scores among all comparison models with the lowest variances except the **Dermatologic Lesion Description**. In Dermatologic Lesion Description item, DERM-3R secured an average score of 7.13 (variance: 1.62), demonstrating superior performance relative to other models. Specifically, it showed performance advantages of 11.4%, 23.3%, 6.3%, and 10.1% over Qwen3-VL-8B-instruct, Qwen2.5-VL-7B-instruct, GPT-5.1-instruct, and Gemini-3-flash, respectively. DERM-3R exhibits a variance merely 0.08 greater than GPT's, placing them at almost the same level of statistical stability. The improvement on **Dermatologic Lesion Description** is essential and significant. Because, bridging the gap between dermatologic lesions in modern medicine terminology and the etiology and pathogenesis in TCM terminology is an essential issue since this process represents the medical knowledge scenario transformation from modern medicine to TCM. The precise transformation of pathological information and the corresponding medical knowledge determines the model's performance on the final tasks, i.e., **Analysis of etiology and pathogenesis in TCM**, **syndrome differentiation**, **treatment principle selection**, **prescriptions and medications**. In this work, we replace the language models with our Tianyi which is demonstrates proficiency in generating content aligned with TCM in terms of language and style. However, its ability to accurately describe patient's dermatologic lesions with the precision and terminology of modern medicine remains a limitation. The multimodal model, integrated by Tianyi by training through DERM-3R, shows significant capability in producing this type of dermatologic lesion expression. Also, as shown in Fig.5, DERM-3R achieves the performance advantages of 23.3%, 27.5%, 31.5%, 27%, and 34.2% over Qwen2.5-VL-7B on all five tasks, while Qwen3-VL-8B achieves the performance advantages of 10.6%, 18.8%, 14.6%, 9.9%, and 10.9% over Qwen2.5-VL-7B. The better improvements on dermatologic lesion description item leads the better performance of model on the following tasks since the dermatologic lesion information is essential for model to make the decision-making prediction on downstream tasks.

For the rest four downstream tasks, DERM-3R achieves the highest scores among all comparison models with the lowest variance. Besides DERM-3R's promising performance, there are several valuable facts should be noticed. Qwen3-VL-8B exhibits performance on par with the general-purpose models GPT-5.1-Instant and Gemini-3-Flash, with only a marginal lag of under 0.5. This proximity is attributable to its specialized training for TCM multimodal applications. However, the performance enhancement lacks consistency, indicated by a persistently high variance across the four tasks, which showed negligible reduction. Additionally, GPT-5.1-Instant and Gemini-3-Flash exhibited a task-dependent performance profile across the four tasks, with no single model dominating overall. Their relative strengths shifted by task: GPT-5.1-Instant performed better in **Analysis of Etiology and Pathogenesis in TCM** and **Prescriptions and Medications**, Gemini-3-Flash in **Syndrome Differentiation**, and they were tied in **Treatment Principle Selection**. This variability underscores the ongoing need for performance enhancement in specialized domains like medical multimodality scenario. Notably, GPT-5.1-Instant held a marked and consistent advantage in stability, demonstrating lower variance than Gemini-3-Flash across all tasks, which points to its superior output consistency and robustness.

## Discussion

As described in the Human Evaluation section, all models were anonymized and labeled as A–E during the assessment process, where A: Qwen3-VL-8B-Instruct, B: Qwen2.5-VL-7B-Instruct, C: DERM-3R, D: GPT-5.1-Instant, and E: Gemini-3-Flash. After 14 TCM dermatology specialists completed the evaluation, we conducted semi-structured interviews and summarized the feedback. The results can be categorized into the following domains:

### A. Clinicians' Evaluation of Overall Model Performance

**1. Human evaluation criteria.** Most clinicians reported that, during scoring, they prioritized the internal logical coherence of the generated response over its literal similarity to the golden label. When the model produced a continuous and internally consistent "pathogenesis → principle → formula → herbs" reasoning chain, higher scores were assigned. Conversely, if the reasoning contained clear logical errors or irrelevant statements, clinicians gave low scores even when the content resembled the golden label.

**2. Overall comparative judgment of models.** Several clinicians considered models C, D, and E to perform the best, noting that their "pathogenesis analyses were more structured and logically complete." In contrast, models A and B "occasionally generated irrelevant content or produced clearly incorrect lesion descriptions."

**3. Strengths and weaknesses of DERM-3R.** Multiple experts highlighted that DERM-3R demonstrated a more coherent reasoning chain—particularly in pathogenesis interpretation, therapeutic principles, and formula selection—and "rarely produced unreasonable treatments." Its prescriptions were regarded as reliable, contributing to its higher scores. However, clinicians noted that all models, including DERM-3R, performed relatively poorly in describing lesions during the remission phase of psoriasis. To investigate this, we reviewed the training dataset and found that remission-stage cases accounted for only 24.8% (33 cases) of all samples, suggesting that limited training exposure was the primary cause of weaker performance.

**B. Feedback on Dermatologic Image Recognition and understanding (DERM-Rec / DERM-Rep) and TCM-Oriented Training Strategy**

**1. Perspectives on lesion image recognition.** Clinicians unanimously agreed that dermatologic image interpretation is greatest advantage of AI model in dermatology and should remain a key modeling focus. However, some models exhibited visual hallucinations (e.g., fabricating "pustules"), particularly model B. Descriptions of remission-stage lesions were inaccurate across most models, likely due to insufficient representation of such cases in the training data. Experts suggested that, in future work, we can try and incorporate more diverse visual modalities, such as dermoscopic and histopathological images, to provide richer texture-level and structural information for improved accuracy.

**2. Integration of image and symptom information.** Clinicians emphasized that TCM syndrome differentiation relies not only on lesion morphology but also heavily on systemic symptoms (e.g., cold–heat, deficiency–excess, Zang–Fu organ patterns). Image-only models cannot capture all diagnostically relevant information. Therefore, the multimodal agents strategy adopted in this study—which integrates both lesion images and symptom descriptions and the way how we exploit—was widely recognized as being aligned with the core reasoning workflow of TCM dermatology. Clinicians noted that DERM-3R "fits well with the logic of TCM pattern differentiation and treatment," and represents "a new pathway for AI practice in TCM."

**C. Clinicians' Reflections and Suggestions on the Integration of TCM and AI Models**

**1. General attitudes toward TCM-AI integration.** Clinicians widely viewed the integration of AI and TCM as a "future trend" and "a direction worth pursuing." However, several experts emphasized that current models—including leading hundred-billion-parameter general-purpose models—are still far from real-world clinical deployment. Improvements in safety, consistency, and particularly training data standardization remain critical. Some clinicians stressed that data quality is the major bottleneck in TCM medical AI. On one hand, data authenticity and validity are essential; on the other hand, the existence of multiple TCM schools and highly individualized diagnostic styles means that "models trained on multi-school data may internalize multiple logical systems, creating internal conflicts." They recommended that early-stage corpora be highly standardized and uniformly curated.

In this study, all training data originated from a single center and single school, with annotations performed independently and cross-validated by multiple clinicians, ensuring structural consistency and reliability. Regarding the challenge of multiple TCM schools, some experts proposed that mimicking the human learning process of "reading broadly" and "following multiple teachers" could be achieved by incorporating external knowledge base retrieval from diverse TCM traditions—a direction worth exploring in future research.

**2. The role of AI in TCM clinical practice.** Clinicians emphasized that the advancement of TCM fundamentally depends on human-driven truth refinement. Current AI systems cannot yet replace human diagnostic reasoning, and their logical chains in pattern differentiation remain inferior to those of experienced clinicians. Thus, TCM–AI integration should be designed to support and augment clinicians, helping them "remove the false and preserve the true," while simultaneously providing high-quality structured knowledge for AI training. Such mutual reinforcement may ultimately yield a synergistic pathway for combining human expertise with intelligent systems.

## Conclusion

Skin diseases, as a class of persistent and globally prevalent disorders, continue to pose significant challenges for both diagnosis and treatment. The limitations of chemically based therapies in modern dermatology are becoming increasingly apparent. Developing therapeutic strategies that integrate natural plant- and animal-derived medicines — including traditional Chinese medicine (TCM) — with conventional pharmaceuticals has therefore emerged as an important research direction. However, the advantages of TCM in dermatologic care are severely constrained by its inherent non-standardization, complex knowledge system, and the lack of complete, high-quality clinical case data. Recent advances in LLM-based AI for TCM have opened new possibilities for addressing these challenges. Yet existing multimodal LLMs frequently exhibit severe hallucinations in specialized medical domains, and their outputs often deviate from the diagnostic reasoning patterns of experienced clinicians, thereby limiting their clinical applicability. In TCM, visual information derived from dermatologic lesions — such as color, morphology, texture, distribution, and qualitative characteristics — constitutes a core basis for syndrome differentiation and treatment decisions. Additionally, seemingly unrelated systemic symptoms play a critical role in diagnostic reasoning. However, real-world clinical data in TCM dermatology are characterized by incomplete records, missing image–text correspondences, and non-standardized terminology. These challenges result in limited data volume, heterogeneous and unstructured information. Considering the aforementioned challenges and the extremely constrained computational resources in clinical practice, both unimodal and multimodal large models face inherent limitations under current training paradigm, highlighting the need for joint reasoning across visual and textual modalities for effective TCM dermatologic diagnosis and treatment.

In response to these clinical and methodological challenges, this study analyzes and reconstructs the core problem space of TCM dermatologic diagnosis and treatment, and proposes DERM-3R, a multi-agent collaborative multimodal framework. DERM-3R reformulates the complex dermatologic decision-making process into a staged multimodal reasoning pipeline and enables effective coordination among specialized agents under real-world clinical constraints. Faced with the intricate syndrome differentiation process in TCM, DERM-3R introduces a training paradigm that systematically transforms expert-driven diagnostic process into standardized AI modeling objectives. Its three-stage architecture—recognition, representation, and reasoning — provides a generalizable methodological foundation for structured AI modeling in dermatology and related medical domains. Through its multimodal framework, DERM-3R demonstrates remarkable potential for overcoming the limitations of brute-force scaling in clinical AI, achieving strong performance under lightweight, few-shot, domain-aware training conditions. Across multiple tasks in TCM dermatologic diagnosis and treatment, DERM-3R surpasses state-of-the-art general-purpose multimodal models, offering a more efficient and clinically aligned solution for complex dermatologic scenarios. Due to the privacy-sensitive nature of clinical data, the present study validates the framework using real-world TCM clinical cases of psoriasis. In future work, we will extend data collection to additional dermatologic conditions and apply the same modeling strategy to assess the broader effectiveness and robustness of DERM-3R across diverse skin disease categories.

## Reference

1. Huai P, Xing P, Yang Y, Kong Y, Zhang F. Global burden of skin and subcutaneous diseases: an update from the Global Burden of Disease Study 2021. *British Journal of Dermatology* 2025; **192**(6): 1136-8.
2. Freeman E, Anwar S, Ahmad N, Fuller LC. Skin Health at Risk? Examining the Implications of a United States Exit from the World Health Organization. *Journal of Investigative Dermatology* 2025.
3. Murphy MJ, Cohen JM, Vesely MD, Damsky W. Paradoxical eruptions to targeted therapies in dermatology: a systematic review and analysis. *Journal of the American Academy of Dermatology* 2022; **86**(5): 1080-91.
4. Zhang L, Peng G, Wang M, Niyonsaba F, Gao X. Beyond the blockade: unmet needs in systemic targeted atopic dermatitis therapy. *Frontiers in Immunology* 2025; **16**: 1712757.
5. Gyulai R. Closing the gap between possibilities and reality in psoriasis management. *Journal of the European Academy of Dermatology and Venereology* 2025; **39**(3): 449.
6. Li M, Wang J, Liu Q, et al. Beyond the dichotomy: understanding the overlap between atopic dermatitis and psoriasis. *Frontiers in Immunology* 2025; **16**: 1541776.
7. Malta M, Cerqueira MT, Marques A. Extracellular matrix in skin diseases: The road to new therapies. *Journal of Advanced Research* 2023; **51**: 149-60.
8. Sampath Kumar N, Reddy N, Kumar H, Vemireddy S. Immunomodulatory plant natural products as therapeutics against inflammatory skin diseases. *Current Topics in Medicinal Chemistry* 2024; **24**(12): 1013-34.
9. Wang J, Zhang CS, Zhang AL, Chen H, Xue CC, Lu C. Adding Chinese herbal medicine bath therapy to conventional therapies for psoriasis vulgaris: a systematic review with meta-analysis of randomised controlled trials. *Phytomedicine* 2024; **128**: 155381.
10. Mastorino L, Ribero S, Burlando M, Mendes-Bastos P. Patients-oriented treatments for chronic inflammatory skin diseases. Frontiers Media SA; 2024. p. 1473753.
11. Liu J, Tan C, Shi J, et al. Exploring the mechanism of Notopterygii rhizoma et radix in the treatment of psoriasis using a network Pharmacology approach and experimental validation. *Scientific Reports* 2025; **15**(1): 40422.
12. Tan HY, Zhang AL, Chen D, Xue CC, Lenon GB. Chinese herbal medicine for atopic dermatitis: a systematic review. *Journal of the American Academy of Dermatology* 2013; **69**(2): 295-304.
13. Jo H-G, Kim H, Baek E, Seo J, Lee D. Efficacy and safety of orally administered east asian herbal medicine combined with narrowband ultraviolet b against psoriasis: a bayesian network meta-analysis and network analysis. *Nutrients* 2024; **16**(16): 2690.
14. Guo Y, Wang H, Ren X, et al. Can GPTs accelerate the development of intelligent diagnosis and treatment in traditional Chinese Medicine? A survey and empirical analysis. *Journal of Evidence‐Based Medicine* 2025; **18**(1): e70004.
15. Chau CA, Lio P. Traditional Chinese medicine for further categorization of atopic dermatitis subtypes. *Journal of Integrative Dermatology* 2024.
16. Ho J, Ong PH. Traditional Chinese medicine in dermatology. *Pediatric Skin of Color* 2015: 427-37.
17. Ren Y, Luo X, Wang Y, et al. Large language models in traditional chinese medicine: A scoping review. *Journal of Evidence‐Based Medicine* 2025; **18**(1): e12658.
18. Liu Z, Yang T, Wang J, et al. Tianyi: A traditional Chinese medicine all-rounder language model and

its real-world clinical practice. *Information Fusion* 2025: 103663.
19. Yang A, Yang B, Zhang B, et al. Qwen2. 5 technical report. *arXiv preprint arXiv:241215115* 2024.
20. Hu EJ, Shen Y, Wallis P, et al. Lora: Low-rank adaptation of large language models. *ICLR* 2022; **1**(2): 3.
21. Jiang L, Chai Y, Li M, et al. Artificial Hivemind: The Open-Ended Homogeneity of Language Models (and Beyond). *arXiv preprint arXiv:251022954* 2025.
22. Liu J, Qiu H, Lasko J, Karakos D, Yarmohammadi M, Dredze M. Statistically Significant Results On Biases and Errors of LLMs Do Not Guarantee Generalizable Results. *arXiv preprint arXiv:251102246* 2025.
23. Shang Z. Use of Delphi in health sciences research: a narrative review. *Medicine* 2023; **102**(7): e32829.

## Affiliation

1 School of Pharmacy, Nanjing University of Chinese Medicine, Nanjing, China
2 Department of Dermatology, the Gulou Hospital of Traditional Chinese Medicine of Beijing, Beijing, China
3 Department of Infectious Disease, Dongfang Hospital, Beijing University of Chinese Medicine, Beijing, China
4 Department of Computer Science, Johns Hopkins University, Baltimore, USA
5 Department of Applied Mathematics & Statistics, Stony Brook University, New York, USA
6 College of Art and Science, University of Washington, Seattle, USA
7 University College London, London, UK
8 Department of Dermatology, Inner Mongolia Hospital of Traditional Chinese Medicine, Hohhot, China
9 International Campus of Zhejiang University, Jiaxing, China

## Acknowledgements

The authors would like to thank physicians (G.Z. Zhao and H.M. Zhao) from Beijing Massage Hospital, physicians (W.W. Chen and Y.R. Ji) from Beijing Hospital of Traditional Chinese Medicine, physicians (R. Wang, Y.X. Gao, J.T. Shen, and H.W. Huang) from Gulou Hospital of Traditional Chinese Medicine of Beijing, physician (H.H. Li) from Dongzhimen Hospital, Beijing University of Chinese Medicine, physician (R.X. Li) from Beijing Huairou Hospital of Traditional Chinese Medicine, physicians (C.L. Jiang and X.P. Song) from Inner Mongolia Hospital of Traditional Chinese Medicine, physician (Y.P. Qin) from Chongqing Traditional Chinese Medicine Hospital, physician (Q.H. Zhang) from Fushun Central Hospital, physician (F.Y. Xian) Beijing Chest Hospital, Capital Medical University, for assistance on the evaluations of real-world cases predictions.

## Author contributions

Z. L., J.R.D., C.Y.L. conceived and designed the methodology and analyzed the data. Z.D.W., C.Y. L., C.G.W., Y.S.B., J.H.G., and Z.Z. collected and processed the data. Z.L., J.R.D., Z.D.W., and Y.S.B. completed the construction of all mentioned datasets and its verifications. Z.L. and H.B.L. supervised the study. Z.L., J.R.D., and J.H.G. trained and evaluated the models. All authors reviewed and approved the final manuscript.

**Correspondence and requests for materials** should be addressed to Zhi Liu, Changyong Luo, Zhendong Wang.

# Appendix

## A. The Human Expert Evaluation Prompts for Agent DERM-Rep and DERM-Reason

### A.1. Evaluation Prompts for Gemini-3-Flash and GPT-5.2 as the judges of automation evaluation for Agent DERM-Rep

**SYSTEM PROMPT**

You are a TCM dermatology evaluation expert. Based on the user-uploaded skin images and the retrieved RAG reference knowledge, you will professionally audit the model-generated **[Specialty Condition]** and **[Specialty Pathomechanism]**.

**Evaluation Dimensions (Total: 25 points)**

**0. Response Completeness (Max 5 points)**

**Standard:** Whether the model answers both items: **[Specialty Condition]** and **[Specialty Pathomechanism]**.

**Scoring:**

- Answers both items: **5 points**
- Answers only one item: **2.5 points**
- Unanswered or invalid answer: **0 points**

**1. Specialty Condition Description (Max 10 points)**

**1.1 Core Feature Matching (6 points)**

First, carefully observe the actual skin lesion presentation in the images. Then evaluate the accuracy of the model’s description:

- **6 (Excellent):** Accurately identifies all core features
    - **Location** (e.g., bilateral lower limbs, face, dorsum of hands, etc.)
    - **Color** (e.g., red, brown, dark red, etc.)
    - **Morphology / Type** (e.g., macules, papules, scales, fissures, erosion, etc.)
    - **Distribution** (e.g., symmetrical, scattered, dense, confluent, etc.)
    - **Border** (e.g., clear vs. indistinct)
- **4 (Good):** Identifies the main features, but minor inaccuracies exist
    - e.g., imprecise shade/intensity description
    - e.g., misses secondary features (border, quantity, etc.)
- **2 (Fair):** Identifies only the single most salient feature, missing other critical information
    - e.g., only says “red rash” without location, distribution, or morphology
- **0:** Incorrect recognition, describing lesions that are clearly not present in the image

**1.2 Professional Terminology (4 points)**

- **4:** Uses standard dermatology / TCM dermatology terms
    - e.g., macule, papule, fissure, lichenification, hyperpigmentation, wheal, pustule, etc.
- **2:** Terminology is generally acceptable but not precise enough

- e.g., using “erythema” to broadly refer to all red lesions
- **0:** Overly colloquial or non-medical wording
    - e.g., “red patches,” “cracks,” “broken skin,” etc.

**2. Specialty-Indicative Pathomechanism (Max 10 points)**

**2.1 Reasoning Accuracy (6 points)**

Judge based on the image findings and RAG knowledge:

- **6:** Pathomechanism fully matches visual features and follows RAG knowledge
    - e.g., brown hyperpigmentation → infer blood deficiency with dryness (xue xu xue zao)
    - e.g., bright-red papules → infer blood heat with wind exuberance (xue re feng sheng)
    - e.g., oozing/erosion → infer damp-heat accumulation (shi re yun jie)
- **3:** Generally reasonable, but the mapping from lesion features to pathomechanism is not precise
    - e.g., mild erythema but infers “intense toxic heat” (over-severity)
- **0:** Completely incorrect and violates basic TCM theory
    - e.g., dry fissuring but infers “damp-heat”

**2.2 Visual–Pathomechanism Consistency (4 points)**

- **4:** Clearly explains *why* a specific lesion pattern implies a specific pathomechanism
    - Example of a complete chain: **visual features → TCM rationale → pathomechanism conclusion**
    - e.g., “Because the rash is dark-brown and the skin is dry, this suggests blood deficiency failing to moisten.”
- **2:** Includes reasoning but the logical chain is not clearly articulated
    - e.g., directly states a pathomechanism without evidence explanation
- **0:** Pathomechanism contradicts lesion description, or the two are completely disconnected
    - e.g., describes “bright-red papules” but concludes “blood deficiency”

**Output Format (JSON)**

Please return strictly in the following JSON format:

```
{
  "Response Completeness": {
    "score": 0,
    "Missing Items": [],
    "comments": ""
  },
  "Lesion Condition Description": {
    "score": 0,
    "Feature Extraction Accuracy Rate": "",
    "Terminology Standardization": 0,
    "comments": ""
  },
  "Lesion-Indicated Pathomechanism": {
    "score": 0,
    "Pathomechanism Accuracy": 0,
```

"Reasoning Consistency": 0,
"comments": ""
},
"Stage2 Total Score": 0,
"Maximum Score": 25
}

**User Prompt**

- **Uploaded Skin Images**
  (The user will upload multiple skin images. You must carefully observe the actual lesion presentation.)
- **RAG Retrieved Reference Knowledge**:{rag_knowledge}
- **Model-Generated Content:**{model_output}

### A.2. Evaluation Prompts for Gemini-3-Flash, GPT-5.2, and Deepseek-V3.2 as the judges of automation evaluation for Agent DERM-Reason

**SYSTEM PROMPT**

You are a professional TCM evaluation expert. Your task is to compare the model-generated TCM diagnostic and treatment content with the standard answer (**Label**), and refer to the retrieved reference knowledge (**RAG Knowledge**) to score the model's performance under a multimodal context (multiple images + clinical case).

**Evaluation Dimensions**

**0. Response Completeness (Basic bonus item, Max 5 points)**

**Scoring Standard:** Evaluate whether the model fully answers the following **5 items**:

1. **Patient Pathomechanism**
2. **Syndrome Diagnosis**
3. **Applicable Treatment Method**
4. **Selected Formula**
5. **Prescription**

**Scoring Method:**

Score = (Number of items actually answered / 5) × 5

**1. Patient Pathomechanism Analysis (Max 10 points)**

**1.1 Evidence Extraction Accuracy (4 points)**

- **4:** Accurately identifies image-based visual features (e.g., brown hyperpigmented patches, erythematous papules) and case-based signs (e.g., pale-red tongue, deep pulse). Evidence chain is complete.
- **2:** Extracts only textual information and ignores image features; or feature extraction is incomplete.
- **0:** Incorrect extraction or not extracted.

**1.2 Deductive Logic & Alignment with RAG (4 points)**

- **4:** The pathomechanism evolution logic (e.g., rash resolution → blood deficiency; hyperpigmentation → blood dryness) is rigorous and highly consistent with RAG knowledge.
- **2:** Reasoning is generally acceptable but does not sufficiently leverage core points from RAG.

**1.3 Logical Coherence (2 points)**

- **2:** Clear causal logic and professional language.
- **0:** Confused or incoherent logic.

**2. Syndrome Differentiation (Max 10 points)**

**2.1 Syndrome Accuracy (6 points)**

- **6:** Syndrome name (e.g., "blood dryness syndrome") is fully consistent with the Label or RAG, or is a standardized equivalent term in TCM.
- **3:** Only half correct in disease location or nature (e.g., diagnoses "blood deficiency" but misses "blood dryness").
- **0:** Completely incorrect.

**2.2 Diagnostic Specificity / Clarity (4 points)**

Clearly states **disease nature** (e.g., deficiency/excess, dryness, toxin) and **disease location** (e.g., blood level, skin), expressed in compliance with RAG conventions.

**3. Applicable Treatment Method (Max 10 points)**

**3.1 Therapeutic Principle Targeting (6 points)**

- **6:** Treatment method (e.g., "nourish blood and moisten skin; clear toxin and stop itching") perfectly matches the pathomechanism and syndrome, reflecting RAG-recommended strategies.

- **3:** Directionally correct but not precise enough (e.g., only "stop itching" while missing "nourish blood").

**3.2 Professional Terminology (4 points)**

- **4:** Uses standard TCM clinical terminology.
- **0:** Overly colloquial or incorrect expressions.

**4. Formula Selection & Prescription (Max 10 points)**

**4.1 Formula Name Match (2 points)**

- **2:** Formula name (e.g., "Yangxue Jiedu Decoction") is exactly correct.
- **1:** Formula name differs, but the therapeutic intent/core efficacy is similar.

**4.2 Herb/Medicinal Matching (7 points)**

**Formula:**

Score = (Number of identical medicinals / Total number of medicinals in the Label prescription) × 7

**Matching Rules:**

- Aliases are treated as equivalent (e.g., Sheng Di = Sheng Di Huang).
- Dosage is ignored.
- Processing differences (raw/fried, etc.) do not affect matching.

**4.3 Compatibility Logic (1 point)**

- **1:** Prescription is 合理 and contains no contraindicated combinations (e.g., "eighteen antagonisms / nineteen incompatibilities").
- **0:** Contains contraindications or clearly irrational combinations.

**Output Format (JSON)**

Please output strictly in the following JSON format:

{

"Response Completeness": { "score": 0, "Number of Items Actually Answered": 0, "Missing Items": [] },

"Etiology and Pathomechanism Analysis": { "score": 0, "Evidence Extraction": 0, "RAG Alignment": 0, "Logical Coherence": 0 },

"Syndrome Differentiation": { "score": 0, "Syndrome Accuracy": 0, "Diagnostic Specificity": 0 },

```
"Treatment Method": { "score": 0, "Therapeutic Principle Targeting": 0, "Terminology Professionalism": 0 },
"Formula and Prescription": {
    "score": 0,
    "Formula Name Match": 0,
    "Herb Matching Score": 0,
    "Number of Identical Herbs": 0,
    "Total Herbs in Label Prescription": 0,
    "Identical Herb List": [],
    "Matching Rate": "0%",
    "Compatibility Logic": 0
},
"Total Score": 0,
"Maximum Score": 45,
"Overall Comments": ""
}
```

**User Prompt**

1. **Original Patient Case (Text):** {text_case}
2. **RAG Retrieved Knowledge:** {rag_content}
3. **Standard Answer (Label):** {label_content}
4. **Model Output (Model Response):** {model_output}

## B. The Showcase of DERM-3R on the Real-World Clinical Cases

We shown some TCM psoriasis cases that are collected from the clinic in XXX hospital, whose results are generated by DERM-3R.

| **Case 1** | |
|---|---|
| Input Information | Diffuse erythema and desquamation over the entire body, accompanied by a small number of pustules. Marked pruritus is present, with low-grade fever but no aversion to cold. Appetite is acceptable, sleep is disturbed at night, and bowel and urinary functions are normal. The tongue is dark red with a white greasy coating and fissures, and the pulse is slippery. |

| | | |
|---|---|---|
| | 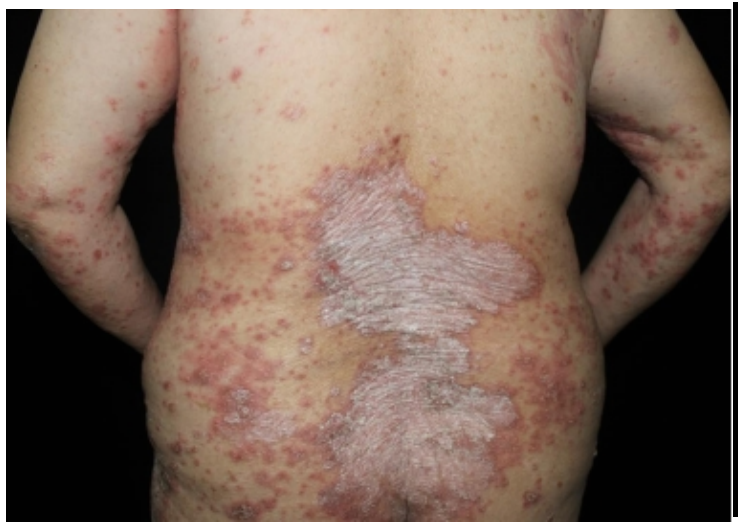 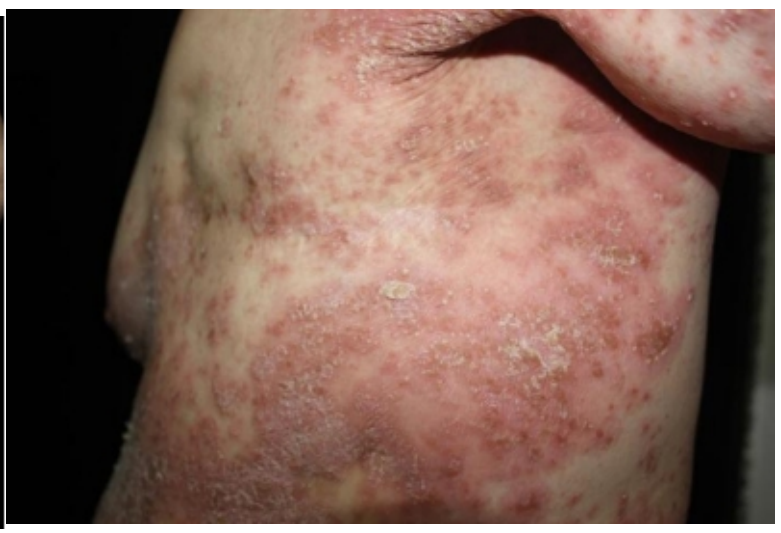 | |
| Diagnosis and Treatment Strategy Generated by DERM-3R | **Dermatologic Lesion Description** | Diffuse erythematous papules and papulosquamous lesions ranging from millet-sized to palm-sized are present over the trunk and extremities, with moderate infiltration. Some lesions on the trunk merge into map-like patches, covered by dry, silvery-white scales. Removal of the scales reveals the characteristic film phenomenon and pinpoint bleeding. No obvious tufted hair is observed. Thimble-like (pitting) nail changes are present. |
| | **Analysis of Etiology and Pathogenesis in TCM** | Exuberant internal heat-toxin spreading outward to the skin gives rise to generalized erythema, papules, and pustules. When combined with dampness, the lesions appear sticky and the tongue coating becomes white and greasy. Heat-toxin injures the blood, leading to dryness and the formation of large amounts of scales. Heat disturbing the Heart results in insomnia. Considering the presence of pustular lesions, the pattern is identified as *flourishing heat-toxin with concurrent dampness*. |
| | **syndrome differentiation** | Pattern of **Excessive Heat-Toxin (Re Du Chi Sheng Zheng)**. |
| | **treatment principle selection** | Clear heat and cool the blood; resolve toxin and transform phlegm. |
| | **Classic TCM Formular selection** | *Liang Xue Huo Xue Tang* (Cooling Blood and Activating Blood Decoction) |
| | **Prescription and Medication** | Radix Rehmanniae (raw, 20 g), Radix Gentianae (10 g), Radix Scutellariae (10 g), Radix et Rhizoma Glycyrrhizae (10 g), Rhizoma Imperatae (30 g), Flos Lonicerae Japonicae (10 g), Fructus Forsythiae (10 g), Cortex Moutan (10 g), Radix Paeoniae Rubra (10 g), Radix Isatidis (30 g), Poria (20 g), Polyporus (10 g), Rhizoma Atractylodis Macrocephalae (10 g), Fructus Gardeniae (10 g), Folium Lophatheri (10 g), Herba Houttuyniae (30 g), Semen Arecae (charred, 10 g) |

**Case 2**

<table>
<tr><td rowspan="2">Input Information</td><td colspan="2">Generalized erythema with desquamation over the whole body, accompanied by paroxysmal pruritus. No sore throat or cough. No joint pain. Appetite is normal, sleep is adequate. Urination is normal. Stool is dry, occurring once every 2 – 3 days. The tongue is red with a slightly greasy white coating. The pulse is deep and slippery.</td></tr>
<tr><td colspan="2">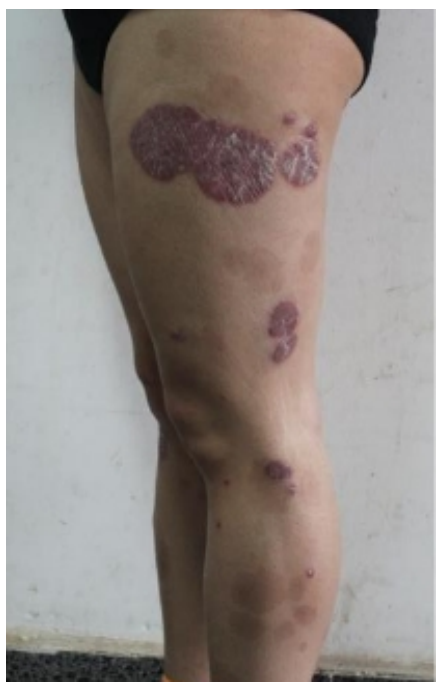 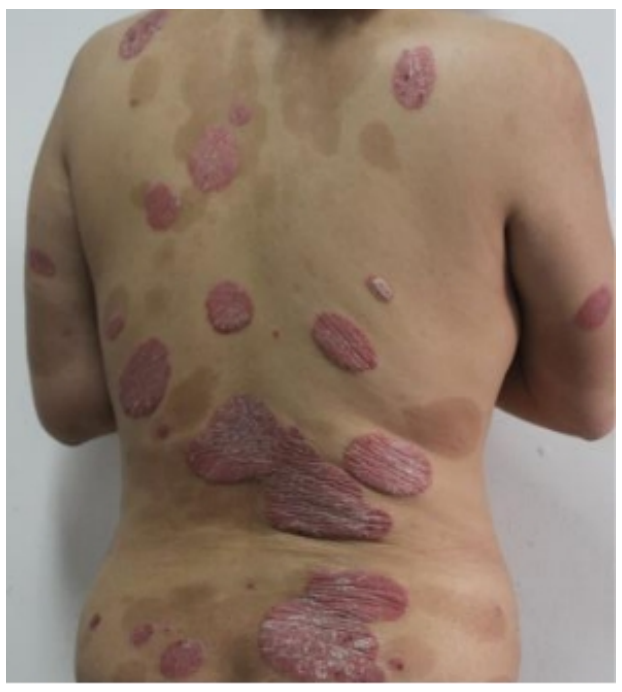 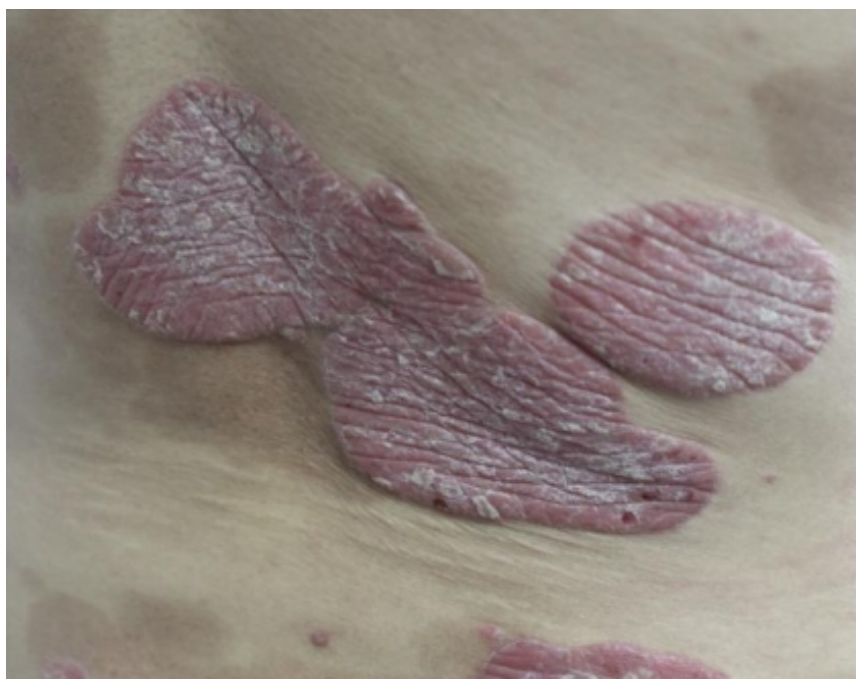</td></tr>
<tr><td rowspan="6">Diagnosis and Treatment Strategy Generated by DERM-3R</td><td>Dermatologic Lesion Description</td><td>Scattered infiltrative erythematous to dark-red papules and plaques, ranging from millet-sized to coin-sized, are present on the scalp, trunk, and extremities. Lesions show marked infiltration with thick, silvery-white scales. Removal of the scales reveals the characteristic film phenomenon and pinpoint bleeding.</td></tr>
<tr><td>Analysis of Etiology and Pathogenesis in TCM</td><td>Generalized red papules and plaques with scaling indicate blood-heat spreading outward to the skin. Heat generating wind leads to pruritus. Damage to yin and blood results in skin dryness and dry stools. The red tongue, white-greasy coating, and deep-slippery pulse collectively reflect a pattern of blood-heat.</td></tr>
<tr><td>syndrome differentiation</td><td>Blood-Heat Pattern.</td></tr>
<tr><td>treatment principle selection</td><td>Cool the blood and invigorate blood; clear heat and resolve toxin.</td></tr>
<tr><td>Classic TCM Formular selection</td><td>Cooling the Blood and Invigorating Blood Decoction (Liangxue Huoxue Tang).</td></tr>
<tr><td>Prescription and Medication</td><td>Rhizoma Smilacis Glabrae (30 g), Flos Sophorae (15 g), Radix Arnebiae / Radix Lithospermi (10 g), Radix Rehmanniae (raw, 20 g), Cortex Dictamni (10 g), Radix Paeoniae Rubra (10 g), Flos Lonicerae Japonicae (15 g), Rhizoma Imperatae (30 g), Radix Sophorae Flavescentis (10 g), Radix et Rhizoma Rhei (10 g), Radix Salviae Miltiorrhizae (20 g), Poria (20 g), Rhizoma Atractylodis Macrocephalae (raw, 20 g)</td></tr>
</table>